\documentclass{article}

\usepackage[preprint,nonatbib]{neurips_2023}
\usepackage{graphicx}
\usepackage{listings}
\usepackage{xcolor}
\usepackage{caption}
\usepackage{subcaption}
\usepackage{xspace}
\usepackage{hyperref}
\usepackage{booktabs}
\usepackage{colortbl}

\hypersetup{
  colorlinks=true, % 设置为 true 以使用颜色
  linkcolor=blue,  % 设置链接颜色为蓝色
  urlcolor=blue    % 设置 URL 链接颜色为蓝色
}
\usepackage[utf8]{inputenc} % allow utf-8 input
\usepackage[T1]{fontenc}    % use 8-bit T1 fontshttps://www.overleaf.com/project/644030e08e064277ba367bae
\usepackage{url}            % simple URL typesetting
\usepackage{booktabs}       % professional-quality tables
\usepackage{amsfonts}       % blackboard math symbols
\usepackage{nicefrac}       % compact symbols for 1/2, etc.
\usepackage{microtype}      % microtypography
\usepackage{xcolor}         % colors
\newcommand{\model}{EmbodiedGPT\xspace}
\newcommand{\dataset}{EgoCOT\xspace}

\title{EmbodiedGPT: Vision-Language Pre-Training via Embodied Chain of Thought}

\lstdefinestyle{bluecode}{
    language=Python, % 代码语言
    basicstyle=\ttfamily\small, % 代码字体样式
    keywordstyle=\color{blue}, % 关键字颜色
    stringstyle=\color{red}, % 字符串颜色
    showstringspaces=false, % 不显示字符串中的空格
    frame=single, % 添加边框
    frameround=tttt, % 边框角落圆形
    framesep=5pt, % 边框与代码之间的距离
    rulecolor=\color{black}, % 边框颜色
    breaklines=true, % 自动换行
    breakatwhitespace=true, % 只在空格处换行
    tabsize=4, % tab键空格数
    numbers=left, % 行号位置
    numberstyle=\small\color{gray}, % 行号字体样式
    stepnumber=1, % 行号间隔
    numbersep=8pt, % 行号与代码之间的距离
    captionpos=b, % 标题位置
    linewidth=0.8\textwidth,
}

\author{%
\textbf{Yao Mu$^1$, Qinglong Zhang$^{2}$, Mengkang Hu$^{1}$, Wenhai Wang$^{2}$, Mingyu Ding\thanks{Corresponding authors: Mingyu Ding and Ping Luo (\{dingmyu, pluo.lhi\}@gmail.com)}~$^{\ ,1}$, Jun Jin$^{3}$,} \\
\textbf{Bin Wang$^{3}$, Jifeng Dai$^{2}$, Yu Qiao$^{2}$, Ping Luo$^{*,1,2}$}\\
$^1$The University of Hong Kong,
$^2$Shanghai AI Laboratory,
$^3$Noah's Ark Laboratory\\
}
\begin{document}

\maketitle

\begin{abstract}
% Embodied AI represents a critical frontier in robotics, where the capacity to devise coherent action sequences for robots to accomplish specific tasks in physical environments is both essential and fraught with inherent challenges. Recently, there has been an increasing interest in developing large foundation models capable of generating embodied plans by processing natural language instructions and environmental observations.

Embodied AI is a crucial frontier in robotics, capable of planning and executing action sequences for robots to accomplish long-horizon tasks in physical environments.
%
% In this work, we introduce \model, an end-to-end multi-modal foundation model for embodied AI, empowering embodied agents with  natural language understanding and execution capabilities.
In this work, we introduce \model, an end-to-end multi-modal foundation model for embodied AI, empowering embodied agents with multi-modal understanding and execution capabilities.
To achieve this, we have made the following efforts:
(i) We craft a large-scale embodied planning dataset, termed \dataset. The dataset consists of carefully selected videos from the Ego4D dataset, along with corresponding high-quality language instructions. Specifically, we generate a sequence of sub-goals with the "Chain of Thoughts" mode for effective embodied planning.
% Specifically, we generate a sequence of sub-goals derived from visual observations with multi-modal "Chain of Thoughts" for effective embodied planning.
%
(ii) We introduce an efficient training approach to \model for high-quality plan generation, by adapting a 7B large language model (LLM) to the \dataset dataset via prefix tuning.
(iii) We introduce a paradigm for extracting task-related features from LLM-generated planning queries to form a closed loop between high-level planning and low-level control. 
% (iii) We created a large-scale dataset of embodied planning, called \dataset. The dataset consists of carefully selected videos from the Ego4D dataset, along with corresponding high-quality language instructions. Specifically, we generated a sequence of sub-goals for planning, derived from visual observations and a multimodal "Chain of Thought".
%
Extensive experiments show the effectiveness of \model on embodied tasks, including embodied planning, embodied control, visual captioning, and visual question answering.
% wwh: 有没有一个让人一眼能记住的指标？
% \textcolor{red}{
Notably, \model significantly enhances the success rate of the embodied control task by extracting more effective features. It has achieved a remarkable 1.6 times increase in success rate on the Franka Kitchen benchmark and a 1.3 times increase on the Meta-World benchmark, compared to the BLIP-2 baseline fine-tuned with the Ego4D dataset. 
\end{abstract}

\section{Introduction}

Embodied AI tasks, e.g., embodied planning, embodied VQA, and embodied control, aim to imbue robots with the ability to perceive, reason, and act within their environment, enabling them to perform long-horizon plans and execute actions autonomously based on real-time observations.
Recently, large language models (LLMs) such as GPT4~\cite{OpenAI2023gpt4} and PaLM-E~\cite{driess2023palme}, have shown promising language understanding, reasoning, and "chain-of-thought" capabilities.
Such advances may open new possibilities for developing robots capable of processing natural language instructions, performing multi-modal "chain-of-thought", and planning actions in physical environments.

Large-scale datasets play important roles in training large language models.
For example, OpenCLIP trains its ViT-G/14 model on the LAION-2B dataset ~\cite{ilharco_gabriel_2021_5143773}, which contains 2B image-language pairs.
Unlike general-purpose visual language tasks that can get a huge amount of weakly labeled image-caption pairs from the Internet, embodied AI tasks require egocentric data in robotics domains.
Also, structured language instructions are needed for precise planning, which usually requires huge manual efforts and costs.
This poses a challenging problem in collecting high-quality embodied multi-modal data.
Some researchers ~\cite{jiang2022vima,zhao2022opend,shridhar2022cliport,zheng2022vlmbench} explore creating large-scale embodied datasets with simulators, but a significant gap remains between simulation and the real world. 
Recent works~\cite{zhang2023llama,gao2023llama,hu2023llm} also explore adapting the pre-trained LLMs to a new domain by efficient tuning strategies like LoRA~\cite{hu2021lora}.
However, several open questions still remain: how to apply LLMs to the field of robotics which may face large domain gaps; how to leverage the "chain-of-thought" capability for structured planning; and how to use the output language plan for downstream manipulation tasks in an end-to-end manner.

\begin{figure}[t]
\vspace{-20pt}
\centering
\includegraphics[width=0.99\linewidth]{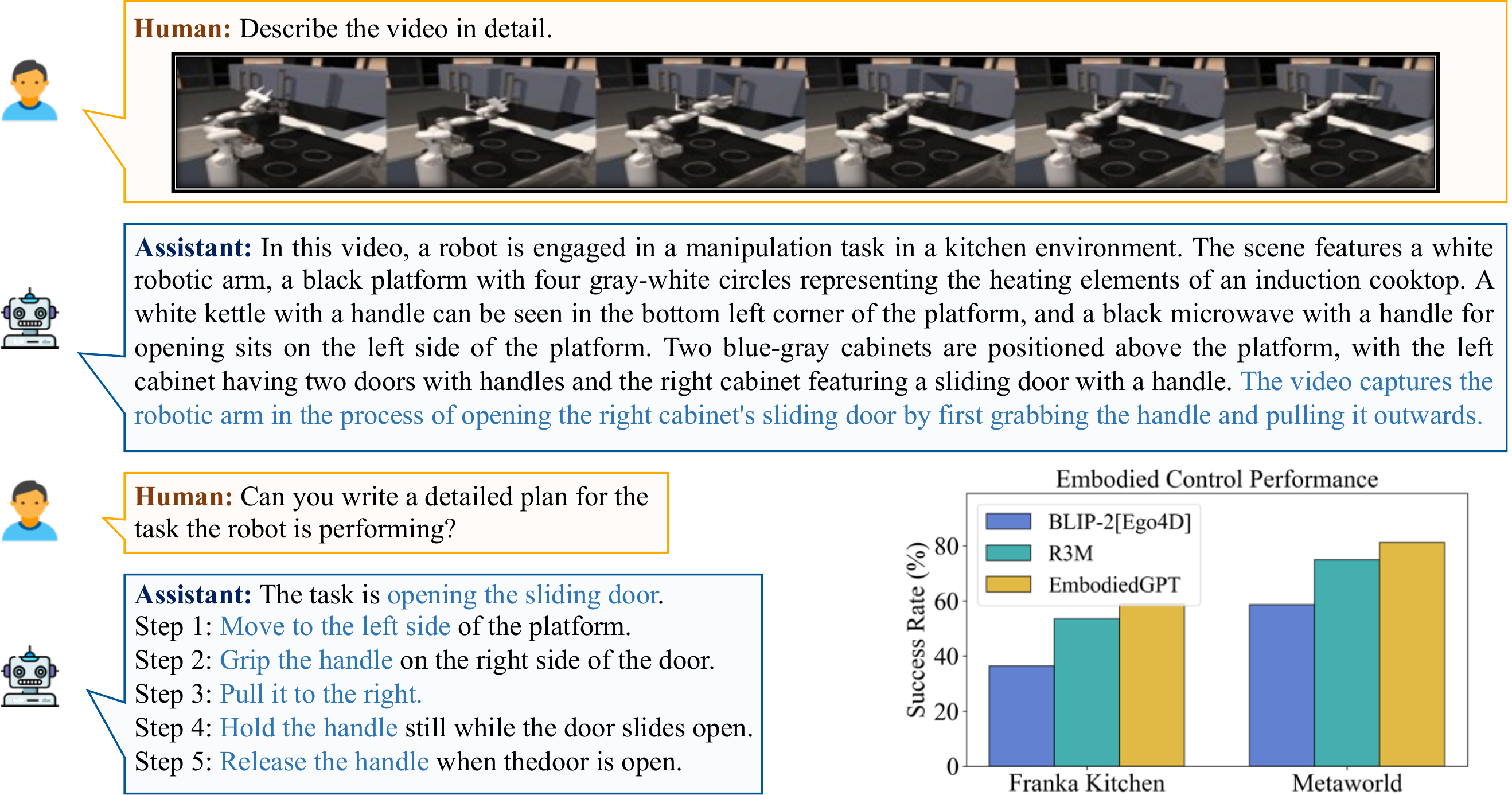}
\caption{EmbodiedGPT's capabilities for video captioning, multi-turn question answering, embodied planning, and low-level control. The plans given by \model are highly executable and incorporate task-specific features, leading to a significant improvement in the success rate of embodied control tasks, outperforming both R3M~\cite{nair2022r3m} (a video-language contrastive learned model) and BLIP-2~\cite{blip2} (a multi-modal foundation model) on Franka Kitchen ~\cite{gupta2019relay} and Meta-World~\cite{yu2020meta} environments.}
 % (a video-language contrastive learned model) 
 % the video-language contrastive learned
 % the multimodal foundation model
\vspace{-12pt}
\label{fig:main_fig}
\end{figure}

To solve the above challenges, in this work, we first build a large-scale embodied planning dataset, termed \dataset, which features chain-of-thought planning instructions.
It contains carefully selected egocentric videos from the Ego4D dataset~\cite{grauman2022ego4d} and corresponding high-quality step-by-step language instructions, which are machine-generated, then semantics-based filtered, and finally human-verified.
Additionally, we also create the EgoVQA dataset as an extension of the Ego4D dataset, focusing on egocentric human-object interaction video question-answering tasks, which aims to offer a wider range of egocentric multi-modal data. 
% Additionally, we develop EgoVQA, an egocentric video question-and-answer dataset based on our \dataset. 
% %
% Enabling the "chain-of-thought" nature of \dataset allows us to formulate the structured plans in \dataset into QA descriptions in EgoVQA, improving the model’s capacity to perform both general and embodied VQA tasks effectively.

Based on our \dataset and EgoVQA, we present an end-to-end multi-modal embodied foundation model called \model, which can interact with the physical world in a more natural and intuitive manner, and perform many embodied tasks, as shown in Figure \ref{fig:main_fig}, such as embodied planning, embodied VQA, and embodied control. 
\model comprises four integrated modules that work together, including i) a  frozen vision model for encoding visual features of current observations, ii) a frozen  language model used to execute natural language for question answering, captioning, and embodied planning tasks, iii) an embodied-former with a language mapping layer for aligning the visual and embodied instructions and extracting task-relevant instance-level features with the generated planning for low-level control, and iv) a policy network, which is responsible for producing low-level actions based on the task-relevant features, enabling the agent to effectively interact with the environment. To further enhance \model’s performance in generating reliable planning containing sub-goal sequences, we implement prefix tuning on the frozen language model to encourage the generation of more executable planning. 

Our method possesses the following core advantages: i) the generated planning exhibits strong executability and granularity at the object part level, such as the gripper of a robotic arm or the handle of a door, manifested in sub-goal sequences. ii) the proposed EgoCOT dataset is built based on an open-source large-scale dataset, which offers greater scalability compared to the PaLM-E~\cite{driess2023palme} model trained on proprietary robot data. And both the EgoCOT dataset, and the \model model will be open-sourced. 
iii) \model forms a closed-loop from high-level planning to low-level control, which enables seamless integration of high-level planning and low-level control, providing efficient task performance and adaptability to a wide range of tasks. To achieve this, we utilize the embodied-former to query task-relevant instance-level features through cross-attention between visual observations and generated embodied planning. This enables the policy network to complete low-level control tasks with fewer than 25 demonstrations.

The contributions can be summarized as follows:
(i) We build an end-to-end multi-modal foundation model \model for embodied AI, which is featured with "chain-of-thought" capability, empowering embodied agents to interact with the physical world in a more natural and intuitive manner.
(ii) We develop two datasets, EgoCOT and EgoVQA, consisting of 200M annotated videos from the Ego4D dataset with corresponding detailed planning instructions and VQA data. The datasets are first machine-generated, then semantics-based filtered, and finally human-verified for quality control.
(iii) We introduce \model a cost-effective training approach and a paradigm for extracting task-relevant features from LLM-generated planning queries, thereby forming a closed loop between high-level planning and low-level control. 
We demonstrate our approach’s effectiveness by achieving state-of-the-art or comparable performance on multiple embodied tasks, including embodied control, embodied planning, video captioning, and video QA. Notably, in comparison to BLIP-2~\cite{li2023blip} fine-tuned on the Ego4D dataset and R3M~\cite{nair2022r3m} specifically designed for manipulation tasks, \model outperforms both models on the Franka Kitchen ~\cite{gupta2019relay} benchmark with a margin of 22.1\% and 5.5\%, respectively. Similarly, on the Meta-World ~\cite{gupta2019relay}  benchmark, \model surpasses both models with margins of 22.5\% and 4.2\%, respectively.

\section{Related Work}
\subsection{Vision Language Pre-training with large scale foundation model}
Vision-Language Pre-training focuses on strengthening the link between visual observation  and natural language. The goal is to develop models that can better understand and process visual content, such as recognizing objects and actions, and generating descriptive text. 
% To better adapt to a range of tasks, different model structures and pre-training objectives have been developed. These include the dual-encoder~\cite{clip,align}, fusion-encoder~\cite{LXMERT,ALBEF}, encoder-decoder~\cite{VL_T5,simvlm,pali}, and unified transformer models~\cite{blip,beit3}. 
% Several pre-training goals are suggested to boost performance in diverse downstream tasks, such as contrasting image-text learning~\cite{clip,filip,ALBEF,blip}, image-text matching~\cite{ALBEF,blip,VLMo}, and (masked) language modeling~\cite{ALBEF,blip,coca,beit3}.
As models become larger, the computational expense for end-to-end pre-training rises, leading to the need for modular vision-language pre-training methods. These methods smartly use pre-trained models, keeping them ‘frozen’ during vision language pre-training to save on computational costs. For example, models like Uniter~\cite{uniter}, Oscar~\cite{oscar}, VinVL~\cite{vinvl}, and LiT~\cite{LiT} freeze the image encoder, while Frozen~\cite{Frozen} and VGPT~\cite{vgpt} freeze the language model. Furthermore, Flamingo~\cite{flamingo} and BLIP-2 ~\cite{li2023blip} use both frozen image encoders and language models, providing a balance between performance and computational efficiency.
% Despite the big steps forward in this area, current models often struggle with complex image-based inference and planning. 
Due to the lack of open-source data for multi-modal embodied planning, previous works struggled to perform detailed task decomposition and lacked the ability to generate precise and executable plans.
% Due to the lack of training data for multi-modal embodied planning, we are typically limited to macro-level planning, making it difficult to perform detailed task decomposition and lacking the ability to generate precise and executable plans.
To tackle this issue, we create the EgoCOT dataset and develop an embodied chain-of-thought vision language pre-training framework to enhance the capacity of multi-modal models for embodied reasoning and planning.

\subsection{Egocentric Video Datasets.}
Egocentric videos, which are captured using wearable cameras, provide a natural perspective of daily activities and pose several challenging research questions~\cite{caba2015activitynet, abu2018will, wong2022assistq}. Several egocentric video datasets have been created over the years, including \cite{damen2022rescaling, sigurdsson2018charades, li2015delving}. However, the collection of egocentric videos is expensive, and previous datasets tend to be small-scale and domain-specific. Recently, a massive egocentric video dataset, Ego4D~\cite{grauman2022ego4d}, has been released and has been used for embodied representation learning. The dataset comprises 3,670 hours of videos collected by 931 people from 74 locations across 9 countries, with videos accompanied by narrations. For embodied AI tasks, learning from large and diverse egocentric human videos has emerged as a promising approach to acquiring a generally useful visual representation for controlling such tasks.  For example, R3M~\cite{nair2022r3m} developed a sparse and compact visual representation using the Ego4D human video dataset through a combination of time-contrastive learning and video-language alignment. VIP~\cite{ma2022vip}, learns general-purpose reward functions for goal-conditioned robotic manipulation using the Ego4D dataset.

\subsection{Large Foundation Model Assistant System}
Recent advancements in large-scale multi-modal language models (LLMs), such as GPT-3~\cite{NEURIPS2020_1457c0d6} and GPT-4~\cite{OpenAI2023gpt4}, have resulted in the creation of various models that can understand multiple modes of information. Two main approaches are used in this field: systematic collaboration and end-to-end trained models.
Systematic collaboration approaches involve coordinating multiple vision models or tools with language models to combine visual information with textual descriptions. Examples include models like Visual ChatGPT~\cite{visualchatgpt}, MM-REACT~\cite{yang2023mm}, and HuggingGPT~\cite{hugginggpt}. However, this approach is limited by the accuracy and capacity of fixed modular models, which can lead to an accumulation of errors. On the other hand, end-to-end models aim to provide unified models for multi-modal tasks. For example, Flamingo~\cite{flamingo} combines vision and language by freezing pre-trained vision encoders and language models. 
BLIP-2~\cite{blip2} introduces Q-Former to align visual features from frozen visual encoders with large language models. Recently, models such as MiniGPT-4~\cite{minigpt4} and LLaVA~\cite{llava} align instruction-tuned language models  with visual features from frozen visual backbones.  VideoChat\cite{li2023videochat}, mPLUG-Owl~\cite{ye2023mplug} and X-LLM~\cite{chen2023x}, further expand support for video input. 
PaLM-E~\cite{palm-e} is the first large embodied multi-modal model, which directly incorporates features from sensor modalities to improve real-world performance and is trained with their large-scale everyday robot data~\cite{ahn2022can}. Compared to PaLM-E, \model is more compact, with a size of only 10B and offers additional support for video captioning, video QA and making planning according to a demonstration video. Furthermore, we form a closed-loop system that spans from high-level planning to low-level control.

% On the other hand, end-to-end models aim to provide unified models for multimodal tasks. For example, Flamingo~\cite{flamingo} combines vision and language by freezing pre-trained vision encoders and language models, demonstrating impressive capabilities in handling new tasks with minimal training. BLIP-2~\cite{li2023blip} introduces Q-Former to align visual features from frozen visual encoders with large language models, while Palm-E~\cite{driess2023palme} directly incorporates features from sensor modalities to improve real-world performance.
% Instruction-tuned language models like Alpaca~\cite{alpaca} and Vicuna~\cite{vicuna}, built upon the LLaMA foundation model~\cite{touvron2023llama}, have shown comparable performance to ChatGPT and GPT-4. Models such as MiniGPT-4~\cite{minigpt4} and LLaVA~\cite{liu2023visual} align these fine-tuned models with visual features from frozen visual backbones. VideoChat\cite{li2023videochat}, mPLUG-Owl~\cite{ye2023mplug} and X-LLM, further expand support for video input. These models represent a significant advancement in the development of multimodal language models, but they have not yet focused on the embodied AI field. One key difference in embodied AI is its emphasis on egocentric input images and videos. Moreover, the interaction between agents and their environments takes center stage in this field, which necessitates agents to comprehend spatial relationships, create clear plans, and execute low-level actions effectively.

\section{Method}
\begin{figure}[t]
\centering
\includegraphics[width=0.99\linewidth]{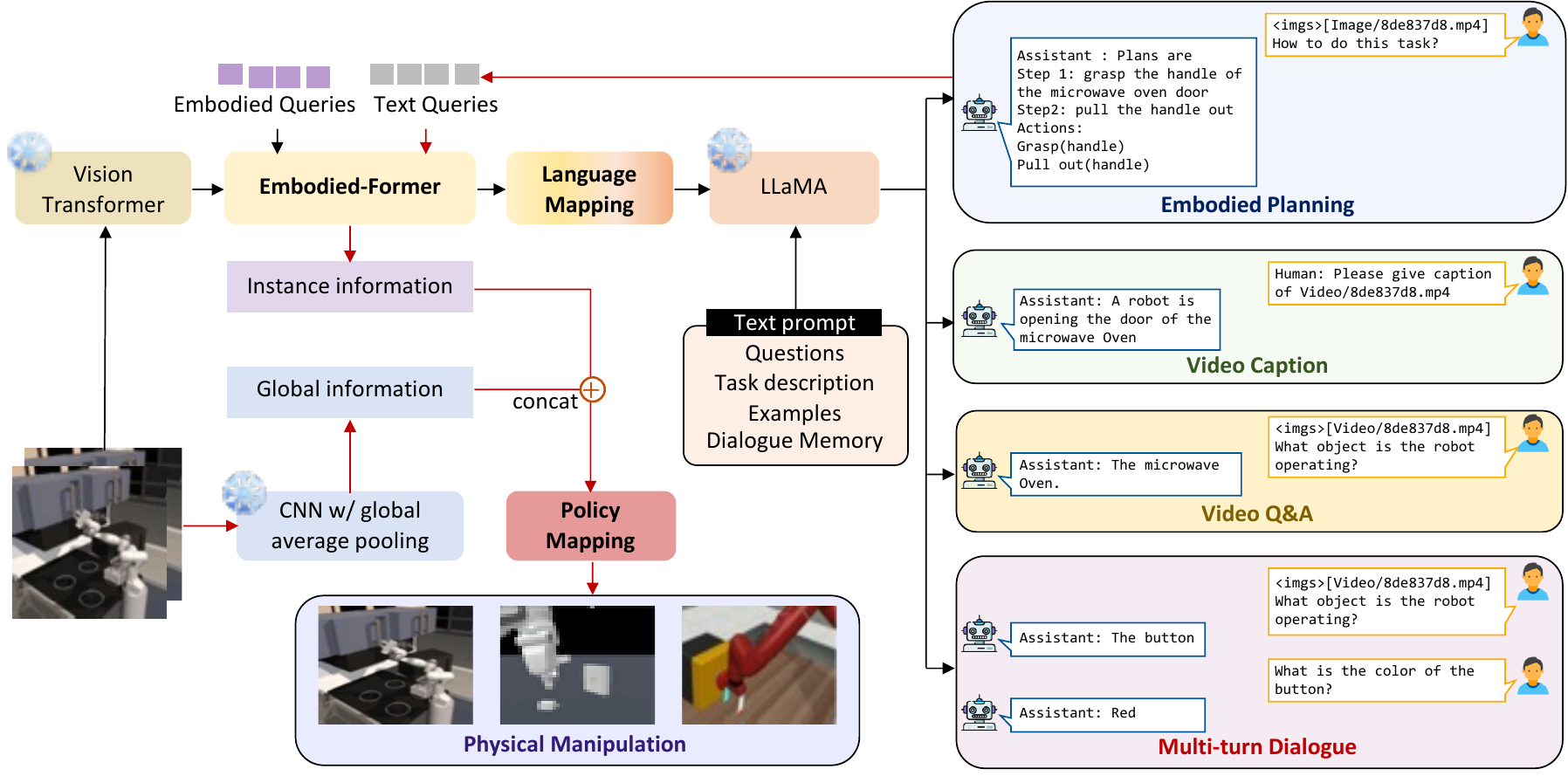}
\caption{Overall framework of \model. The black arrow shows the vision-language planning process, while the red arrow represents that we leverage the queried language plans for better policy learning in low-level control tasks.}
\vspace{-15pt}
\label{fig1:Overall framework}
\end{figure}

% The key to constructing a unified embodied foundation model is to develop a visual feature extraction backbone that can support concept comprehension tasks, such as video captioning and video question answering (VQA), while also being easily adaptable to low-level motion generation for embodied control tasks. Our approach achieves this goal by creating a novel video-language pre-training paradigm that uses a chain of thought to generate task plans from egocentric video inputs.
% The chain of thought generation task is a unique type of video Q\&A task that asks "how to complete the task." To implement this, we use a simple prefix text, "how to do the task that + \textit{original caption}". This task not only improves video Q\&A performance but also encourages the visual extractor to learn features that are specific to task completion, making them more suitable for embodied control tasks. This is a crucial innovation in our paper.
% To separate the extraction of general visual features and task-specific features, we also employ a prompt tuning mechanism. This mechanism cleverly avoids interference between concept comprehension tasks and low-level control tasks. For example, background information may be necessary for VQA tasks but can impede learning efficiency and robustness in low-level control tasks. By designing learnable query tokens and their interaction with visual tokens, we can effectively avoid these problems. Downstream tasks only require low-cost fine-tuning of query tokens.

The goal of the embodied foundation model is to imitate human-like perception and interaction with the environment by accurately perceiving the environment, identifying relevant objects, analyzing their spatial relationships, and formulating a detailed task plan. 
To achieve this, the EmbodiedGPT employs a pre-trained vision transformer as the visual encoder and a pre-trained LLaMA~\cite{touvron2023llama}  model as the language model. As shown in Figure \ref{fig1:Overall framework}, the embodied-former acts as a bridge between the visual and language domains, it first extracts compact visual features from the output of the vision model through attention-based interaction involving visual tokens, text queries, and learnable embodied queries and then maps it to the language modality through a language mapping layer.
These embeddings are sent to the frozen LLaMA~\cite{touvron2023llama} language model for visual caption, visual QA, and embodied planning.  
The generated planning is then used to query highly relevant features from the general visual tokens encoded by the visual model via the embodied-former. These features are utilized to generate low-level control commands for task execution through the downstream policy network. To enhance performance across a range of embodied tasks, we introduce a novel video-language pre-training paradigm that leverages a cognitive chain of thought to produce embodied planning from egocentric video inputs. We formulate this task as a standard VQA (Visual Question Answering) task, using "how to do the task that + original caption" as the question and embodied planning as the answer. This framework enriches the data of embodied planning and standard visual QA tasks, encouraging the embodied-former to capture task-specific features that are more suitable for embodied control tasks.

% The embodied foundation model aims to imitate human-like perception and interaction with the environment by accurately perceiving objects, analyzing their spatial relationships, and formulating a detailed task plan.
% the Embodied-former serves as a bridge between the visual and linguistic domains by extracting the final visual encoding through cross-attention with learnable query tokens and mapping it to the language modality through a language mapping layer. 
% The Embodied Foundation Model aims to accurately perceive the environment from an egocentric perspective by selecting relevant objects based on a clear task description, analyzing their spatial relationships, and formulating a comprehensive task plan.
% % To achieve this, improve performance on both diverse embodied tasks and standard visual captioning and question-and-answer tasks
% To achieve this goal, we introduce a novel video-language pre-training paradigm that leverages a cognitive chain of thought to produce task plans from egocentric video inputs. This unique type of video QA task queries "how to complete the task" and is achieved by using a simple prefix text, "how to do the task that + original caption." This framework enriches the data of embodied planning and standard visual QA tasks, encouraging the embodied-former to capture task-specific features that are more suitable for embodied control tasks.

\subsection{Framework}
The training process consists of three stages, each designed to incrementally develop reasoning and planning capabilities. The first two stages focus on pre-training in basic cognitive and responsive skills, while the third stage involves training the embodied AI task with egocentric video-text data on EgoCOT. In the first stage, we focus on image-text conversation alignment pre-training, which involves using three datasets: COCO Caption~\cite{coco}, 595 thousand finely filtered image-text pairs from CC3M~\cite{cc3m}, and 491 thousand filtered image-text pairs obtained by re-captioning LAION-400M using BLIP-2~\cite{li2023blip}. The primary goal of this stage is to pre-train the Embodied-former and language projection while keeping the vision and language model parameters frozen to save computational resources. 
In the second stage, our goal is to enhance the model's ability to comprehend and generate more complex sentences and improve its reasoning skills. We achieve this by updating the language projection and prefix language adapter and utilizing the "Complex\_Reasoning\_77k" and multi-turn conversation datasets provided by "LLaVA\_Instruct\_150K"~\cite{liu2023visual}.
% The second stage is a fine-tuning phase, targeting the language projection and prefix language adapter. Here, we employ the "Complex\_Reasoning\_77k"  and multi-round conversation datasets provided by "LLaVA\_Instruct\_150K" ~\cite{liu2023visual}. This step's focus is on enhancing the model's proficiency in deciphering and generating more sophisticated language structures, fostering its reasoning capabilities.

\textbf{Embodied "chain-of-thought" training with EgoCOT}: During the third stage, we first use Conv3D~\cite{singh20193d} to transfer the pre-trained vision model from stage 2 to the video encoder, with a time offset of 2 and a total frame count of 8 for the videos. Then, we introduce the 'chain-of-thought' vision language pre-training paradigm where the model takes 8 keyframes of the video as input, along with the task description, embodied planning, and structured verb-noun pairs summary to reason with a prompt, such as Listing \ref{lst:pretrain}. To avoid overfitting, we provide a prompt set that has different instructions with the same meaning. In this stage, we fine-tune the patch embedding, the language projection layer, and the prefix language adapter to better capture temporal information.
% \vspace{-10pt}
% In the third stage, we first transfer the pre-trained vision model in stage 2 to the video encoder by employing Conv3D~\cite{singh20193d} for patch embedding, with a time offset of 2 and a total frame count of 8 for the videos. 
% Then we introduce the embodied "chain-of-thought" vision language pre-training paradigm, \model takes the video's 8 key frames as input, and given the task description, embodied planning, and structured verb-noun pairs summary as reasoning with the prompt like Listing \ref{lst:pretrain}, we provide a prompt set which has different instruction with the same meaning to overcome the overfitting problem.  During this stage, we finetune the patch embedding to better capture temporal information, the language projection layer, and the prefix language adapter.

% In the third stage, we fine-tune the model using the EgoCOT dataset, and 

\definecolor{codegreen}{rgb}{0,0.8,0}
\definecolor{codegray}{rgb}{0.5,0.5,0.5}
\definecolor{codepurple}{rgb}{0.58,0,0.82}
\definecolor{backcolour}{rgb}{0.99,0.99,0.97}

\vspace{-12pt}
\begin{lstlisting}[
float=htbp,
language=Python,
floatplacement=htbp,
frame=single,
frameround=tftf,
belowskip=-1\baselineskip,
basicstyle=\ttfamily\scriptsize,
breakatwhitespace=false,
breaklines=true,
captionpos=b,
keepspaces=true,
showspaces=false,
showstringspaces=false,
showtabs=false,
label={lst:pretrain},
caption=Prompt we used for chain-of-thought pre-training.]
Watch this video, identify the actions and devise a plan using chain-of-thought. Extract detailed actions using this schema:
Task: {"task description"}
Plan: {"plan with chain-of-thought"} Actions: {{"number"}: {'verb'}({'noun'})}.
\end{lstlisting}
\vspace{-8pt}

\subsection{Model Architecture}
The Embodied-former, denoted as $\mathcal{E}(\cdot)$, serves as a bridge between visual input 
$x_{\mathrm{vis}}$ and the frozen language model, acting as an information bottleneck that delivers the most relevant visual data to the language model.
The Embodied-former consists of two sub-modules: one for extracting features from the image input, denoted as $\mathcal{E}_{\mathrm{vis}}: x_{\mathrm{vis}} \rightarrow y_{\mathrm{vis}}$, and another for extracting features from the text input, denoted as $\mathcal{E}_{\mathrm{txt}}: x_{\mathrm{txt}} \rightarrow y_{\mathrm{txt}}$.
We employ $N$ learnable embodied query embeddings $y_{\mathrm{query}}$ as the input of $\mathcal{E}$ to interact with $x_{\mathrm{vis}}$ through cross-attention layers and with $x_{\mathrm{txt}}$ through self-attention layers. We denote the output query representation as $z \in \mathbb{R}^{N \times D}$, where $D$ is the dimensionality of the embeddings. The dimension of $z$ is significantly smaller than that of the visual features.
The output query embeddings are then transformed to $z^{'} \in \mathbb{R}^{N \times D^{'}}$, which have the same dimensionality $D^{'}$ as the LLM's text embedding in the language modality. This transformation is performed by a mapping function denoted as $M: z \rightarrow z^{'}$, which is accomplished by a linear projection via a fully-connected (FC) layer. The projected embeddings, $z'$, serve as "soft visual prompts for the language model," decoupling the whole interaction into visual-query interaction and query-text interaction.  The final embodied planning is inferred by the language model with $z'$ and text prompt(shown as Listing \ref{lst:pretrain}) as input. For low-level control which aims to generate actions to interact with the environment, the embodied plan $x_{\mathrm{plan}}$ is used as input text for embodied-former to query the task-relevant instance level features 
$z_{\mathrm{instance}} = \mathcal{E}(x_{\mathrm{vis}},x_{\mathrm{plan}},y_{\mathrm{query}})$. 
Subsequently, the agent is capable of generating control commands, such as the turning angle of the servo, represented as $a = g(z_\mathrm{instance}, z_{\mathrm{global}})$. This function combines both the instance-specific information $z_\mathrm{instance}$ and the global context $z_{\mathrm{global}}$. The global context is inferred using a ResNet50 model~\cite{he2016deep} that has been pre-trained on ImageNet~\cite{deng2009imagenet}, employing global average pooling.
% Subsequently, the agent can generate control commands, for instance, the turning angle of the servo, denoted as $a=g(z_\mathrm{instance},z_{\mathrm{global}})$, which consolidates both the instance-specific information, $z_\mathrm{instance}$, and the global context, $z_{\mathrm{global}}$, which is inferred through a ResNet50~\cite{he2016deep} pre-trained on ImageNet ~\cite{deng2009imagenet} utilizing global average pooling. 
Here, $g(\cdot)$ represents the policy network, which is a Multi-Layer Perceptron (MLP)~
\cite{riedmiller2014multi} mapping function. The output of the policy network consists of specific executable actions, such as positions and velocities in the Cartesian coordinate system. More implementation details can be found in Appendix A.

\subsection{Training Settings}

% \textbf{Pre-Trained Model}:
We employ the same pre-trained image encoder as BLIP-2\cite{li2023blip}. Specifically, we utilize the ViT-G/14 model from EVA-CLIP~\cite{eva} and remove its last layer, using the output features of the second last layer instead. 
For the frozen language model, we adopt a pre-trained LLaMA-7B~\cite{touvron2023llama} model and fine-tune it using the ShareGPT dataset and a GPT-4 generated 52K English instruction-following dataset~\cite{gpt4_ins}. We then utilize the well-fine-tuned language model as the frozen language model for vision-language pre-training.
Additionally, we convert the data type of parameters of the frozen ViT~\cite{dosovitskiy2020image} and language model to FP16 during pre-training to increase efficiency.

% \textbf{Training Settings}:
% We use the same set of training hyper-parameters for all models during vision-language pre-training. We employ the AdamW optimizer~\cite{adamw} with $\beta_1=0.9$, $\beta_2=0.98$, and a weight decay of 0.05. We also utilize a cosine learning rate decay with a peak learning rate of $2\times10^{-5}$ and a linear warm-up of 2 thousand steps. Our training data consists of images of size 224$\times$224 that are augmented with random resized cropping and horizontal flipping.

\subsection{Creating EgoCOT and EgoVQA Dataset}
For our EgoCOT dataset, we obtain basic data from the Ego4D dataset~\cite{grauman2022ego4d}, which includes $9,645$ untrimmed videos of various durations ranging from 5 seconds to 7 hours. 
% EGOVLP \cite{NEURIPS2022_31fb284a} further refined the Ego4D dataset by meticulously cleaning the data and clipping the videos into standard 60-second segments. Each of these segments has been paired with a concise text description providing context for the video content. Building upon EGOVLP's open-source data, we conducted an additional round of cleaning to ensure the quality and consistency of the data. Specifically, we filtered out all videos without human-object interaction, such as watching TV or walking, and removed the videos with too large a gap from indoor robot scenes, such as mixing cement at a construction site.
To prepare the data for our purposes, we conducted two stages of data cleaning to prepare our data. In the first stage, we filtered out videos with missing or very short narrations (which made up 7.4\% and 0.9\% of the text, respectively), as well as those with unsure tags (which accounted for  4.0\% of the text). We also excluded videos without human-object interaction, such as watching TV or walking. After this stage, we were left with 2.9 thousand hours of video, containing 3.85 million narrations, from 129 different scenarios covering 2927 hours of video.

\definecolor{codegreen}{rgb}{0,0.8,0}
\definecolor{codegray}{rgb}{0.5,0.5,0.5}
\definecolor{codepurple}{rgb}{0.58,0,0.82}
\definecolor{backcolour}{rgb}{0.99,0.99,0.97}

% To create pairs of caption, text planning of chain of thought, and video clips with corresponding time intervals, we follow the framework of EgoVLP \cite{NEURIPS2022_31fb284a} to segment the video. Specifically, narrations per video in Ego4D are organized as a sequence of sentences ${\mathcal{T}_0, \cdots, \mathcal{T}_n}$ with exact timestamps ${t_0, \cdots, t_n}$, indicating that an event $i$ described by $\mathcal{T}_i$ happened at the moment $t_i$.
% For a given narration $\mathcal{T}i$ with timestamp $t_i$, we pair a clip $\mathcal{V}i$ with the following start and end timepoints:
% \begin{equation}
% [t_i^{start}, t_i^{end}]=[t_i-\beta_i/2\alpha,~t_i+ \beta_i/2\alpha],
% \label{pairing}
% \end{equation}
% where $\beta_i=\sum_{j=0}^{n-1}\left(t_{j+1}-t_j\right) / n$ is an adjustable parameter equal to the average temporal distance between pairs of consecutive narrations in a given video. Conversely, $\alpha$ is a scale factor computed as the average of all $\beta_i$ across all videos in the EgoCOT~($\alpha=4.9$ seconds).
To generate pairs of captions, embodied plannings, and corresponding video segments with time intervals, we utilized the EgoVLP framework \cite{NEURIPS2022_31fb284a} to segment the video. The narrations are organized as a sequence of sentences ${\mathcal{T}_0, \cdots, \mathcal{T}_n}$ with precise timestamps ${t_0, \cdots, t_n}$ that indicate when a described event occurred.
For each narration $\mathcal{T}_i$ with timestamp $t_i$, we paired it with a clip $\mathcal{V}_i$ by determining its start and end time points:
\begin{equation}
[t_i^{start}, t_i^{end}]=[t_i-\beta_i/2\alpha,~t_i+ \beta_i/2\alpha],
\label{pairing}
\end{equation}
where $\beta_i=\sum_{j=0}^{n-1}\left(t_{j+1}-t_j\right) / n$ is an adjustable parameter equal to the average temporal distance between consecutive narrations in a given video. Conversely, $\alpha$ is a scale factor computed as the average of all $\beta_i$ across all videos in the EgoCOT dataset ($\alpha=4.9$ seconds).
For each video segment, we provide prompts and corresponding captions for ChatGPT~\cite{openai_chatgpt_2023} to generate a reasonable and detailed embodied planning. The caption is typically a brief introduction such as "C opens a drawer."
We use the ChatGPT to generate a chain of thought according to the caption and organize it into a list of verb-noun pairs, such as \textcolor{darkgray!80}{\textbf{\textit{"plans: grasp the handle with the gripper and pull the handle; actions: 1. grasp(handle, gripper) 2. pull(handle)."}}} The prompt we used to generate EgoCOT dataset is shown in Listing \ref{lst:egocot}. To enhance the diversity of generated chain of thoughts, we employ a temperature parameter of 0.9 and a top-p parameter of 0.95. For each prompt, we perform five sampling iterations.

\vspace{-12pt}
\begin{lstlisting}[
float=htbp,
language=Python,
floatplacement=htbp,
frame=single,
frameround=tftf,
belowskip=-1\baselineskip,
basicstyle=\ttfamily\scriptsize,
breakatwhitespace=false,                   
breaklines=true,                 
captionpos=b,                    
keepspaces=true,                            
showspaces=false,                
showstringspaces=false,
showtabs=false,                  
label={lst:egocot},
caption=Prompt we used for creating EgoCOT dataset.]
You need to generate plans with chain of thought for each task, and then extract  detailed actions (collocation of nouns and verbs) from the plan. 
The action can be of the following form:
[action_name], eg., turn left;
[action_name] argument1, eg., pick up(apple);
[action_name] argument1 argument2, eg., put(apple, table)
Task: pick up a cup on the table
plans: grasp the handle of the cup with the gripper and lift it up
Actions:
1. grasp(handle of the cup, gripper)
2. lift up(cup)
\end{lstlisting}

\textbf{Post-procedure.} To ensure the quality of the generated planning instructions, we perform the second stage of  data cleaning. We used the CLIP model \cite{clip} to assess the similarities between the video and text pairs. For each video, we compared it against five potential embodied plans and selected the one with the highest similarity as the corresponding label for the embodied plan.
We then took our data-cleaning process a step further by filtering out any video-caption-planning pairs with similarities lower than the threshold. We eliminated both data with the low similarity between the video and caption and between the video and planning to ensure the highest quality data for our EgoCOT dataset. For each keyframe of the video segment, we use the CLIP model to encode both the text data $T$ and the image data $I$ into a shared embedding space. The similarity is calculated using the cosine similarity function as $S(y_{T},y_{I}) = \frac{y_{T} \cdot y_{I}}{\|y_{T}\|\|y_{I}\|}$, where $S(y_{T},y_{I})$ denotes the similarity between the text and image, and $y_{T}$ and $y_{I}$ are the respective embeddings.
Given that each video contains multiple keyframes, an ensemble of similarity scores is obtained for each video. This ensemble strategy helps to alleviate the problem of variability among individual frames and ensures a more robust and representative measure of overall similarity. The ensemble similarity score between  a video $V$ with $n$ keyframes and text data $T$  is given by:
\begin{equation}
E(V,T) = \frac{1}{n}\sum_{i=1}^{n}S({y_{T}}_{i},{y_{I}}_{i})    
\end{equation}
where $E(V,T)$ is the ensemble similarity score, $S({y_{T}}_{i},{y_{I}}_{i})$ is the similarity score for the $i$-th keyframe, and $n$ is the total number of keyframes.
% We then took it a step further and filtered the video and embodied planning pairs. 
% To ensure the highest data quality and accuracy, we also eliminated all generated embodied plannings with a similarity to the video lower than 0.25. These meticulous filtering steps will undoubtedly contribute to the success of our research endeavors and allow us to make meaningful contributions to the field.
We also created the EgoVQA dataset specifically for egocentric human-object interaction video question-answering tasks to enrich the training data. For each caption in the Ego4D dataset, we used ChatGPT to generate five QA pairs. To ensure relevance, we guided ChatGPT to focus on core key verbs and nouns by designing prompts as shown in Listing \ref{lst:code3}. The sampling schema when crafting EgoVQA is the same to that as EgoCOT.

% We also developed a video QA dataset called EgoVQA for tasks involving egocentric human-object interaction videos. In this dataset, we have integrated the video caption task and the video Q\&A task by designing distinct text prefixes for language model input. When generating video captions, we use the prefix "Describe this video." On the other hand, for video QA, the prefix is left empty. To create question-answering pairs, we rely on ChatGPT's original caption and replace core key verbs and nouns while carefully designing the prompt, as shown in Listing \ref{lst:code3}. This process allows us to generate relevant and accurate question-answering pairs.

\vspace{-12pt}
\begin{lstlisting}[
float=htbp,
language=Python,
floatplacement=htbp,
frame=single,
frameround=tftf,
belowskip=-1\baselineskip,
basicstyle=\ttfamily\scriptsize,
breakatwhitespace=false,                   
breaklines=true,                 
captionpos=b,                    
keepspaces=true,                            
showspaces=false,                
showstringspaces=false,
showtabs=false,                  
label={lst:code3},
caption=Prompt used for creating EgoVQA dataset.]
Please ask some questions accroding to the verbs and nouns in the sentence. 
For example, in this sentence "a man is picking up a cup", the verb is picking up and the noun is cup, therefor questions can be "what is the object the man is picking up?" or "what operation is performed on the cup?". 
Then You need to give the answer.

input: a man is picking up a cup
question: What is the object the man is picking up
answer: The cup
\end{lstlisting}
\vspace{-5pt}

\begin{table}[b]
  \centering
  \small
    \begin{tabular}{lccccc}
    \toprule
    Model   & Object($\uparrow$) & Spatial($\uparrow$) & Redundancy($\downarrow$) & Plan Reasonable($\uparrow$) & Plan Executable($\uparrow$)   \\
    \hline
    Minigpt4 & 5.6      & 4.8   & 4.4   & 4.5   & 4.8   \\
    LLaVA-7B & 7.3      & 7.4   & 3.9   & 7.5   & 6.6   \\
    LLaVA-13B & \textbf{8.5}     & 8.6   & 3.4   & 8.4   & 7.6 \\
    \cellcolor{gray!20}EmbodiedGPT &  \cellcolor{gray!20}{8.4}   & \cellcolor{gray!20}\textbf{8.8}   & \cellcolor{gray!20}\textbf{2.6}   & \cellcolor{gray!20}\textbf{8.8}   & \cellcolor{gray!20}\textbf{8.4}   \\ 
    \toprule
    \end{tabular}
    \caption{Generate Quality Evaluation on image input tasks.}
    \label{tab:user}
 \vspace{-0.3in}
\end{table}
\section{Experiments}
In this section, we present a comprehensive evaluation of multi-modal foundation models and \model, across various tasks including visual captioning, embodied planning, and  control.
% Furthermore, we also show some insight about prompt designing for multi-modal foundation model to improve the relevance of visual input and language generation.

% for concrete tasks (e.g., making a cup of coffee), abstract understanding (e.g., determining appropriate actions when feeling hot), and a comparison with Visual ChatGPT in the Virtual-Home Benchmark. We provide a thorough analysis of the results and a scholarly discussion of the model’s performance in different scenarios.

\begin{figure}[t]
 \vspace{-0.2in}
\centering
 \includegraphics[width=1.0\textwidth]{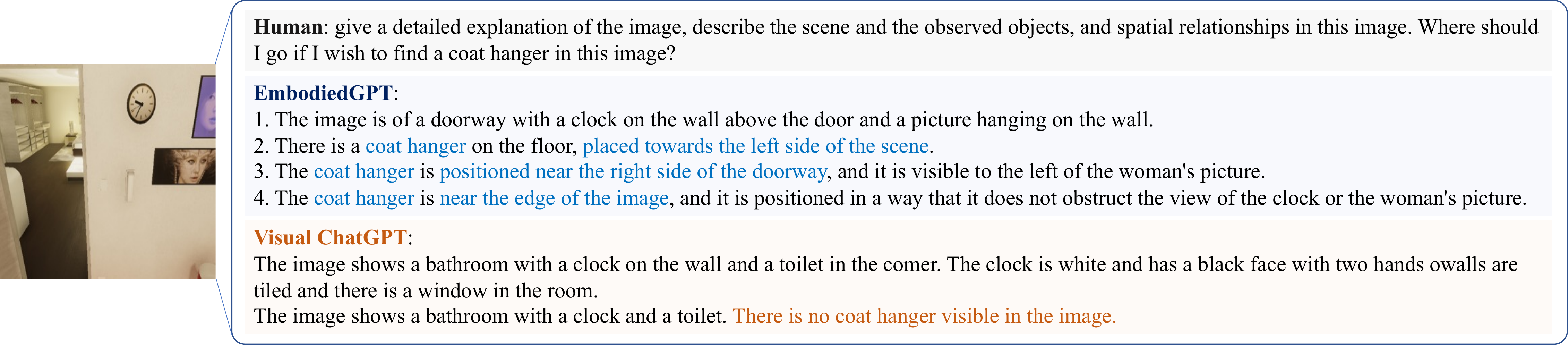}
 \vspace{-0.2in}
    \caption{Comparison between \model and VisualGPT in the question-answering task.}
    \label{fig:compare_visualgpt}
     \vspace{-0.1in}
\end{figure}

\textbf{Evaluation on image input tasks.} In order to evaluate the quality of generated captions and planning with the given image, we conducted a user study with 30 participants. The study included 10 cases of image caption tasks from MS-COCO dataset ~\cite{coco}, 5 embodied planning scenarios in different embodied AI simulators, and 5 real-world scenes with accompanying embodied planning tasks. Participants were asked to rate the generated captions from different end-to-end models on five dimensions using a scoring system ranging from 1 to 10: object recognition accuracy, spatial relationship understanding, level of redundancy in the answer, and reasonability of the planning and the executability of the planning. The average scores among all the participants for different models are shown in Table~\ref{tab:user}.
% The results of the study were analyzed and summarized as shown in Table \ref{tab:user}.  
The results demonstrate that \model achieves a comparable level of object recognition and spatial relationship understanding as the LLaVA-13B model, despite having only 7B parameters in the language model. Furthermore, \model generates less redundant content in relation to the given embodied AI task, and produces the most reasonable and executable planning outputs.
We also compared the performance of \model with Visual ChatGPT~\cite{visualchatgpt}, which adopts a hierarchical approach by combining several pre-trained vision models and language models to answer questions. In the Virtual-Home ~\cite{puig2018virtualhome} benchmark, Visual ChatGPT uses a visual caption model to generate dense captions that are subsequently passed into ChatGPT for deriving a solution. As shown in Figure \ref{fig:compare_visualgpt}, Visual ChatGPT failed to find a coat hanger due to its limitations of relying solely on the caption model for extracting visual information, resulting in poor performance when compared to the end-to-end model like \model. These findings highlight the advantages of adopting a unified, end-to-end model over hierarchical approaches that rely on multiple stages.
\begin{figure}[t]
\centering
\includegraphics[width=0.9\linewidth]{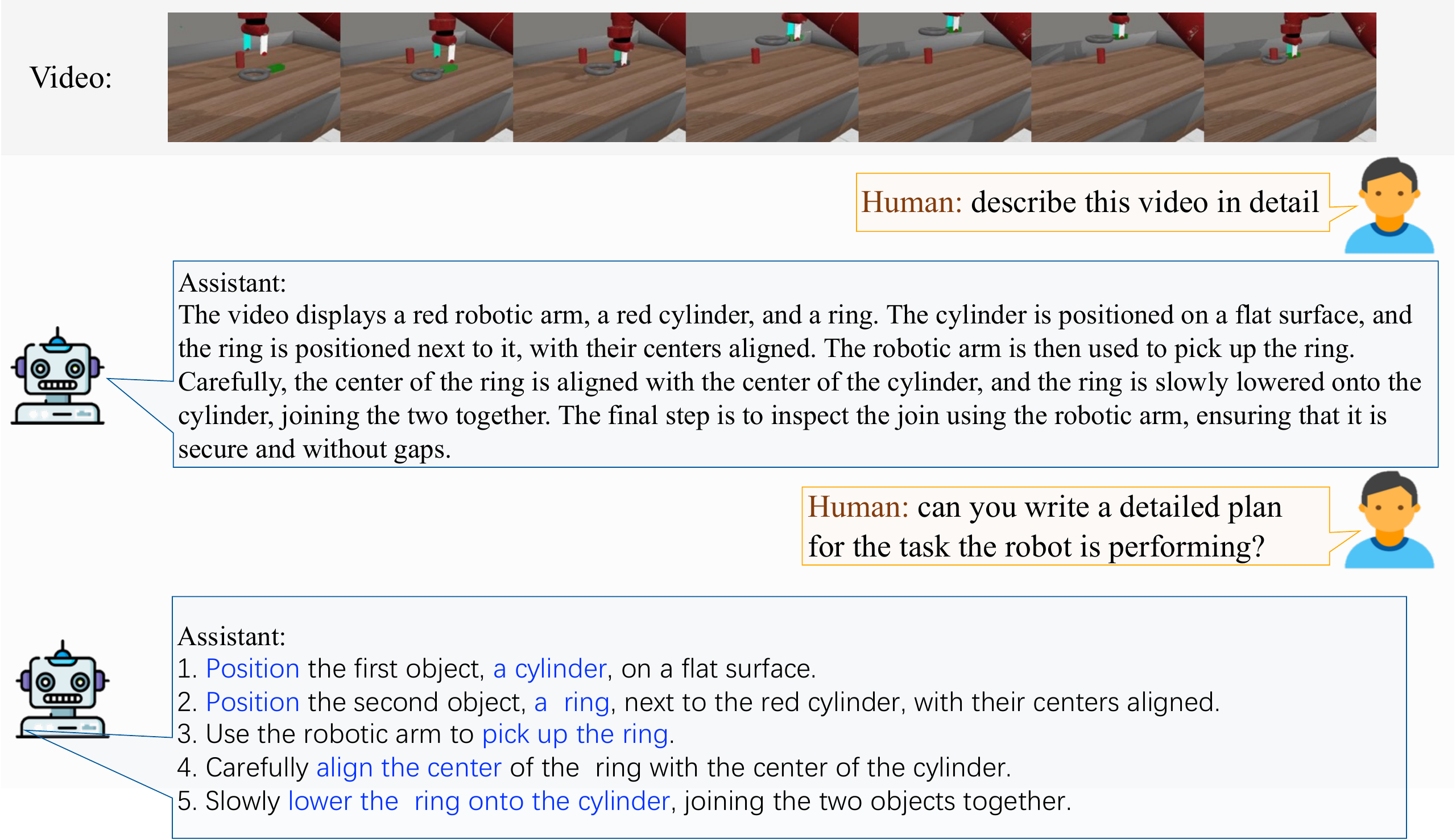}
\vspace{-2pt}
\caption{Example of video input embodied AI tasks on Meta-World benchmark. \model accurately analyzes embodied control tasks in demonstration videos and provides precise planning. }
\vspace{-18pt}
\label{fig:main_fig2}
\end{figure}

\begin{figure}[htbp]
% \vspace{-20pt}
\vspace{-10pt}
    \centering
    \begin{subfigure}[b]{0.98\textwidth}
        \centering        \includegraphics[width=0.99\textwidth]{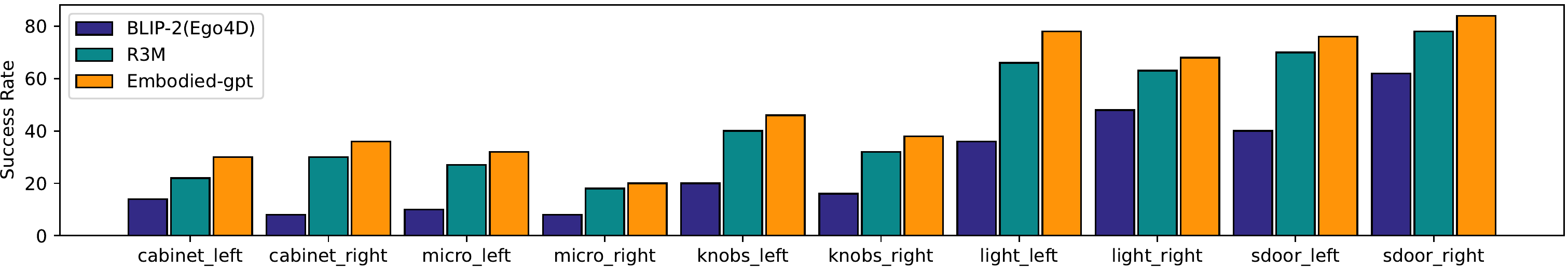}
        \vspace{-0.05in}
        \caption{Performance comparison in \textit{Franka Kitchen} with only 10 demos.}
        \label{fig:y Franka with only 10 demos}
    \end{subfigure}
    \hfill
    \begin{subfigure}[b]{0.98\textwidth}
        \centering
 \includegraphics[width=0.99\textwidth]{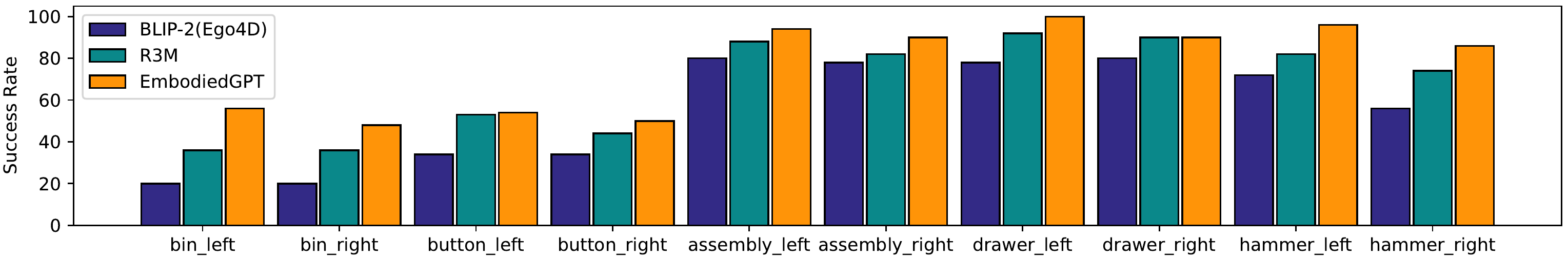}
 \vspace{-0.05in}
        \caption{Performance comparison in \textit{Meta-World} with only 10 demos.}
        \label{fig:meta with 10 demos}
    \end{subfigure}
    \vspace{-2pt}
    \caption{Performance of EmbodiedGPT in low-level control tasks with 10 demonstration demos.}
    \label{fig:10 demos graphs}
\vspace{-6pt}
\end{figure}
\begin{figure}[htbp]
    % \label{fig:25 demos}
    \centering
    \begin{subfigure}[b]{0.98\textwidth}
        \centering      \includegraphics[width=0.99\textwidth]{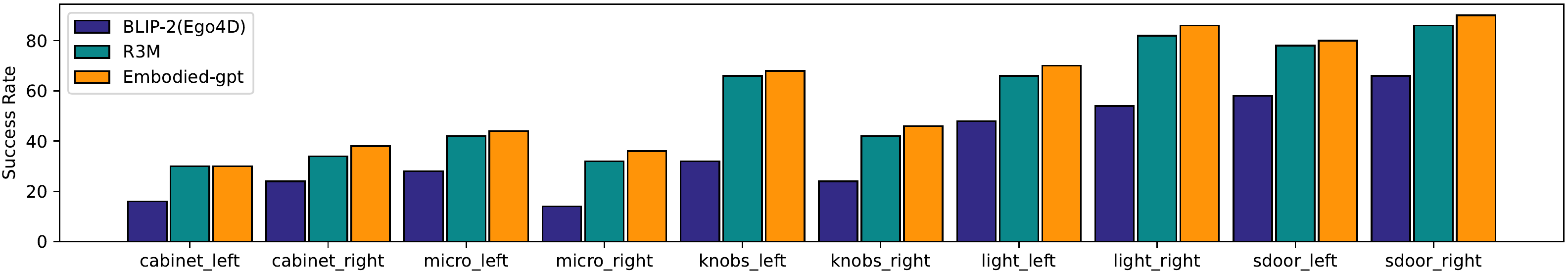}
        \vspace{-0.05in}
        \caption{Performance comparison in \textit{Franka Kitchen} with 25 demos.}
        \label{fig:y equals x}
    \end{subfigure}
    \hfill
    \begin{subfigure}[b]{0.98\textwidth}
        \centering
\includegraphics[width=0.99\textwidth]{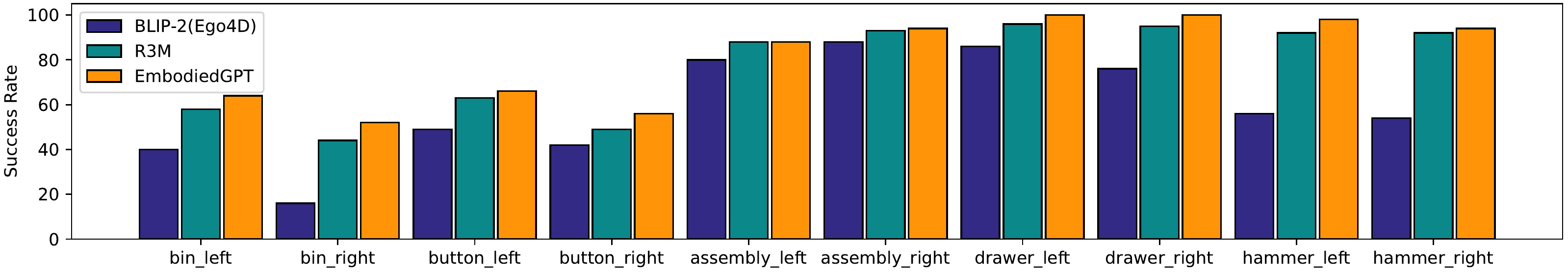}
\vspace{-0.05in}
        \caption{Performance comparison in \textit{Meta-World} with 25 demos.}
        \label{fig:three sin x}
    \end{subfigure}
    \vspace{-2pt}
    \caption{Performance of EmbodiedGPT in low-level control tasks with 25 demonstration demos.}
    \label{fig:25 demos}
\vspace{-10pt}
\end{figure}
\begin{table}[htbp]
\setlength{\tabcolsep}{3.0pt}
\small
\centering
\begin{tabular}{lcccc}
\toprule
Model & Franka(10 demos) & Franka(25 demos) & Meta-World(10 demos) & Meta-World(25 demos) \\
\hline
EmbodiedGPT & \textbf{50.8}\% $\pm2.8$     & \textbf{58.5}\% $\pm2.7$      & \textbf{76.4}\% $\pm2.2$         & \textbf{81.2}\%$\pm2.0$  \\
- Close-loop & 38.6\% $\pm2.9$     & 47.3\% $\pm2.5$     & 62.7\% $\pm2.2$        & 64.9\%  $\pm2.0$       \\
- COT        & 26.2\% $\pm3.2$     & 36.4\% $\pm2.7$     & 55.2\%  $\pm2.4$       & 58.7\% $\pm2.0$   \\
\toprule
\end{tabular}
\caption{Ablation on the closed-loop spans from planning to low-level control, and "chain-of-thought" (COT) training with 25 and 10 demonstrations("-" symbol indicates "removing"). We report the average success rate over 5 tasks and 2 camera views per benchmark. }
\label{tab:ablation}
\vspace{-22pt}
\end{table}

% We are the first to develop a multimodal large-scale model for embodied AI tasks from an ego-centric perspective.
\textbf{Evaluation on video input embodied AI tasks.}  We evaluate the recognition ability of videos and planning abilities of our model for embodied control tasks on standard embodied AI benchmarks, Franka Kitchen~\cite{gupta2019relay} and Meta-World~\cite{yu2020meta}. 
% EmbodiedGPT's understanding 
Meta-World provides a challenging set of tasks that require complex object manipulation skills, including assembling a ring on a peg, picking and placing a block between bins, pushing a button, opening a drawer, and hammering a nail. Franka Kitchen benchmark focuses on tasks like sliding open the right door, opening the cabinet, turning on the light, turning the stovetop knob, and opening the microwave. As shown in Figure \ref{fig:main_fig2}, 
given a demonstration video, \model can accurately interpret the embodied control task and provide step-by-step planning. The output planning is fed into the Embodied-former module of \model to query highly relevant features for use as inputs in the policy network and the low-level  actions are generated  by the policy network to interact with the environment (see more visualizations in Appendix B). 
% We test 100 times for each task and  record their success rates.

\textbf{Evaluation on embodied control tasks.} For embodied control tasks, we compare our model with R3M\cite{nair2022r3m}, which is the state-of-the-art method in these two benchmarks, and an ablation version called 'BLIP-2[Ego4D]', which has the same structure and same amount of parameters as \model,  and is only fine-tuned on the video caption task using the Ego4D dataset without incorporating EgoCOT.
% [Ego4d]". "BLIP-2[Ego4d]" has the same structure and same amount of parameters as \model, finetuned in video caption task with Ego4d dataset, and 
% BLIP-2 model on two benchmarks. 
% The BLIP-2 model has the same transformer structure as the \model and its vision model is finetuned with the Ego4D dataset to ensure fairness in comparison.
In all experiments, the policy network is learned using few-shot learning on a small amount of demonstration data. There are two settings, one of which utilizes 10 demonstrations, and the other utilizes 25 demonstrations. We report the success rate in 100 random evaluations with only visual observations in 5 tasks per benchmark over 5 seeds and 2 different camera views for each setting, respectively. As shown in Figure \ref{fig:10 demos graphs} and Figure \ref{fig:25 demos}, \model outperforms the baseline methods, demonstrating the effectiveness of learning with EgoCOT.

\textbf{Ablation study.}
We perform ablation studies to analyze the effectiveness of the "Chain-of-Thought" training mode and the importance of a closed-loop design for embodied control. 
The results, as shown in Table \ref{tab:ablation}, demonstrate a significant improvement in success rate when using the EgoCOT approach compared to training solely with the EGO4D caption task. Moreover, the closed-loop design is necessary as the generated plans contained specific and relevant sub-goal information, which proved crucial for control tasks.

In summary, \model exhibits a strong ability to generate reasonable planning, accurately extract task-relevant features from visual inputs, as well as execute low-level actions to interact with the environment. The ablation experiments demonstrate that both the training paradigm based on EgoCOT and the closed-loop design from embodied planning to low-level control significantly contribute to the performance improvement of \model.

% In summary, \model exhibits a strong ability to accurately extract relevant features from visual inputs that are most related to the task, and generate reasonable planning outputs as well as interact with the environment to complete embodied tasks. It has achieved excellent performance in visual captioning, visual question answering, embodied planning, and embodied control tasks.

% \vspace{-5pt}
\section{Conclusion}
% \vspace{-5pt}
% In this paper, we propose \model, an end-to-end multi-modal foundation model for embodied AI that enables embodied agents to make step-by-step planning and execute low level commands. Our contributions include the creation of a large-scale embodied planning dataset called \dataset, an efficient training approach for high-quality plan generation using prefix adapters, and the introduction of a paradigm for closed loop high-level planning and low-level control.
% In this paper, we present \model, an end-to-end multi-modal foundation model for embodied AI that empowers agents to perform step-by-step planning and execute low-level commands. To accomplish this, we develop a large-scale embodied planning dataset called \dataset. Additionally, we present an innovative training approach that utilizes prefix adapters to generate high-quality plans with "chain-of-thought". Moreover, we build a embodied control paradigm for closed-loop high-level planning and low-level control establishes seamless coordination between the two levels of decision-making. 
% Through extensive experiments, we have demonstrated the effectiveness of \model on various embodied tasks such as embodied planning, embodied control, visual captioning, and visual question answering, achieving state-of-the-art  or comparable performance. We believe that \model takes a significant step towards developing more intelligent and capable embodied AI agents. 
In this paper, we present \model, an end-to-end multi-modal foundational model for embodied AI that enables agents to perform step-by-step planning and execute low-level commands. To achieve this, we create a large-scale embodied planning dataset called \dataset and develop an efficient training approach that utilizes prefix tuning to generate high-quality plans with a "chain-of-thought". Furthermore, our embodied control paradigm seamlessly coordinates high-level planning and low-level control. Extensive experiments demonstrate the effectiveness of \model on various embodied tasks, achieving state-of-the-art or comparable performance. We believe that \model represents a significant step towards developing more intelligent embodied AI agents.\\
\textbf{Future works and limitations:} \model freezes the parameters of the vision and language model due to limited computational resources. Joint training with all modules and exploring other modalities, such as speech, could be future works. We do not foresee obvious undesirable ethical or social impacts at this moment.
%This paper has no negative social impact.

{\small
\bibliography{main}

\begin{thebibliography}{10}

\bibitem{OpenAI2023gpt4}
OpenAI.
\newblock Gpt-4 technical report, 2023.

\bibitem{driess2023palme}
Danny Driess, Fei Xia, Mehdi S.~M. Sajjadi, Corey Lynch, Aakanksha Chowdhery,
  Brian Ichter, Ayzaan Wahid, Jonathan Tompson, Quan Vuong, Tianhe Yu, Wenlong
  Huang, Yevgen Chebotar, Pierre Sermanet, Daniel Duckworth, Sergey Levine,
  Vincent Vanhoucke, Karol Hausman, Marc Toussaint, Klaus Greff, Andy Zeng,
  Igor Mordatch, and Pete Florence.
\newblock Palm-e: An embodied multimodal language model.
\newblock In {\em arXiv preprint arXiv:2303.03378}, 2023.

\bibitem{ilharco_gabriel_2021_5143773}
Gabriel Ilharco, Mitchell Wortsman, Ross Wightman, Cade Gordon, Nicholas
  Carlini, Rohan Taori, Achal Dave, Vaishaal Shankar, Hongseok Namkoong, John
  Miller, Hannaneh Hajishirzi, Ali Farhadi, and Ludwig Schmidt.
\newblock Openclip, July 2021.

\bibitem{jiang2022vima}
Yunfan Jiang, Agrim Gupta, Zichen Zhang, Guanzhi Wang, Yongqiang Dou, Yanjun
  Chen, Li~Fei-Fei, Anima Anandkumar, Yuke Zhu, and Linxi Fan.
\newblock Vima: General robot manipulation with multimodal prompts.
\newblock {\em arXiv preprint arXiv: Arxiv-2210.03094}, 2022.

\bibitem{zhao2022opend}
Yizhou Zhao, Qiaozi Gao, Liang Qiu, Govind Thattai, and Gaurav~S Sukhatme.
\newblock Opend: A benchmark for language-driven door and drawer opening.
\newblock {\em arXiv preprint arXiv:2212.05211}, 2022.

\bibitem{shridhar2022cliport}
Mohit Shridhar, Lucas Manuelli, and Dieter Fox.
\newblock Cliport: What and where pathways for robotic manipulation.
\newblock In {\em Conference on Robot Learning}, pages 894--906. PMLR, 2022.

\bibitem{zheng2022vlmbench}
Kaizhi Zheng, Xiaotong Chen, Odest~Chadwicke Jenkins, and Xin Wang.
\newblock Vlmbench: A compositional benchmark for vision-and-language
  manipulation.
\newblock {\em Advances in Neural Information Processing Systems}, 35:665--678,
  2022.

\bibitem{zhang2023llama}
Renrui Zhang, Jiaming Han, Aojun Zhou, Xiangfei Hu, Shilin Yan, Pan Lu,
  Hongsheng Li, Peng Gao, and Yu~Qiao.
\newblock Llama-adapter: Efficient fine-tuning of language models with
  zero-init attention.
\newblock {\em arXiv preprint arXiv:2303.16199}, 2023.

\bibitem{gao2023llama}
Peng Gao, Jiaming Han, Renrui Zhang, Ziyi Lin, Shijie Geng, Aojun Zhou, Wei
  Zhang, Pan Lu, Conghui He, Xiangyu Yue, et~al.
\newblock Llama-adapter v2: Parameter-efficient visual instruction model.
\newblock {\em arXiv preprint arXiv:2304.15010}, 2023.

\bibitem{hu2023llm}
Zhiqiang Hu, Yihuai Lan, Lei Wang, Wanyu Xu, Ee-Peng Lim, Roy Ka-Wei Lee,
  Lidong Bing, and Soujanya Poria.
\newblock Llm-adapters: An adapter family for parameter-efficient fine-tuning
  of large language models.
\newblock {\em arXiv preprint arXiv:2304.01933}, 2023.

\bibitem{hu2021lora}
Edward~J Hu, Yelong Shen, Phillip Wallis, Zeyuan Allen-Zhu, Yuanzhi Li, Shean
  Wang, Lu~Wang, and Weizhu Chen.
\newblock Lora: Low-rank adaptation of large language models.
\newblock {\em arXiv preprint arXiv:2106.09685}, 2021.

\bibitem{nair2022r3m}
Suraj Nair, Aravind Rajeswaran, Vikash Kumar, Chelsea Finn, and Abhinav Gupta.
\newblock R3m: A universal visual representation for robot manipulation.
\newblock {\em arXiv preprint arXiv:2203.12601}, 2022.

\bibitem{blip2}
Junnan Li, Dongxu Li, Silvio Savarese, and Steven C.~H. Hoi.
\newblock {BLIP-2:} bootstrapping language-image pre-training with frozen image
  encoders and large language models.
\newblock {\em CoRR}, abs/2301.12597, 2023.

\bibitem{gupta2019relay}
Abhishek Gupta, Vikash Kumar, Corey Lynch, Sergey Levine, and Karol Hausman.
\newblock Relay policy learning: Solving long-horizon tasks via imitation and
  reinforcement learning.
\newblock {\em arXiv preprint arXiv:1910.11956}, 2019.

\bibitem{yu2020meta}
Tianhe Yu, Deirdre Quillen, Zhanpeng He, Ryan Julian, Karol Hausman, Chelsea
  Finn, and Sergey Levine.
\newblock Meta-world: A benchmark and evaluation for multi-task and meta
  reinforcement learning.
\newblock In {\em Conference on robot learning}, pages 1094--1100. PMLR, 2020.

\bibitem{grauman2022ego4d}
Kristen Grauman, Andrew Westbury, Eugene Byrne, Zachary Chavis, Antonino
  Furnari, Rohit Girdhar, Jackson Hamburger, Hao Jiang, Miao Liu, Xingyu Liu,
  et~al.
\newblock Ego4d: Around the world in 3,000 hours of egocentric video.
\newblock In {\em Proceedings of the IEEE/CVF Conference on Computer Vision and
  Pattern Recognition}, pages 18995--19012, 2022.

\bibitem{li2023blip}
Junnan Li, Dongxu Li, Silvio Savarese, and Steven Hoi.
\newblock Blip-2: Bootstrapping language-image pre-training with frozen image
  encoders and large language models.
\newblock {\em arXiv preprint arXiv:2301.12597}, 2023.

\bibitem{uniter}
Yen{-}Chun Chen, Linjie Li, Licheng Yu, Ahmed~El Kholy, Faisal Ahmed, Zhe Gan,
  Yu~Cheng, and Jingjing Liu.
\newblock {UNITER:} universal image-text representation learning.
\newblock In {\em ECCV}, volume 12375, pages 104--120, 2020.

\bibitem{oscar}
Xiujun Li, Xi~Yin, Chunyuan Li, Pengchuan Zhang, Xiaowei Hu, Lei Zhang, Lijuan
  Wang, Houdong Hu, Li~Dong, Furu Wei, Yejin Choi, and Jianfeng Gao.
\newblock Oscar: Object-semantics aligned pre-training for vision-language
  tasks.
\newblock In {\em {ECCV}}, pages 121--137, 2020.

\bibitem{vinvl}
Pengchuan Zhang, Xiujun Li, Xiaowei Hu, Jianwei Yang, Lei Zhang, Lijuan Wang,
  Yejin Choi, and Jianfeng Gao.
\newblock Vinvl: Making visual representations matter in vision-language
  models.
\newblock {\em arXiv preprint arXiv:2101.00529}, 2021.

\bibitem{LiT}
Xiaohua Zhai, Xiao Wang, Basil Mustafa, Andreas Steiner, Daniel Keysers,
  Alexander Kolesnikov, and Lucas Beyer.
\newblock Lit: Zero-shot transfer with locked-image text tuning.
\newblock In {\em {CVPR}}, pages 18102--18112, 2022.

\bibitem{Frozen}
Maria Tsimpoukelli, Jacob Menick, Serkan Cabi, S.~M.~Ali Eslami, Oriol Vinyals,
  and Felix Hill.
\newblock Multimodal few-shot learning with frozen language models.
\newblock In Marc'Aurelio Ranzato, Alina Beygelzimer, Yann~N. Dauphin, Percy
  Liang, and Jennifer~Wortman Vaughan, editors, {\em NeurIPS}, pages 200--212,
  2021.

\bibitem{vgpt}
Jun Chen, Han Guo, Kai Yi, Boyang Li, and Mohamed Elhoseiny.
\newblock Visualgpt: Data-efficient adaptation of pretrained language models
  for image captioning.
\newblock In {\em {CVPR}}, pages 18009--18019, 2022.

\bibitem{flamingo}
Jean{-}Baptiste Alayrac, Jeff Donahue, Pauline Luc, Antoine Miech, Iain Barr,
  Yana Hasson, Karel Lenc, Arthur Mensch, Katie Millican, Malcolm Reynolds,
  Roman Ring, Eliza Rutherford, Serkan Cabi, Tengda Han, Zhitao Gong, Sina
  Samangooei, Marianne Monteiro, Jacob Menick, Sebastian Borgeaud, Andrew
  Brock, Aida Nematzadeh, Sahand Sharifzadeh, Mikolaj Binkowski, Ricardo
  Barreira, Oriol Vinyals, Andrew Zisserman, and Karen Simonyan.
\newblock Flamingo: a visual language model for few-shot learning.
\newblock {\em arXiv preprint arXiv:2204.14198}, 2022.

\bibitem{caba2015activitynet}
Fabian Caba~Heilbron, Victor Escorcia, Bernard Ghanem, and Juan Carlos~Niebles.
\newblock Activitynet: A large-scale video benchmark for human activity
  understanding.
\newblock In {\em CVPR}, pages 961--970, 2015.

\bibitem{abu2018will}
Yazan Abu~Farha, Alexander Richard, and Juergen Gall.
\newblock When will you do what?-anticipating temporal occurrences of
  activities.
\newblock In {\em CVPR}, pages 5343--5352, 2018.

\bibitem{wong2022assistq}
Benita Wong, Joya Chen, You Wu, Stan~Weixian Lei, Dongxing Mao, Difei Gao, and
  Mike~Zheng Shou.
\newblock Assistq: Affordance-centric question-driven task completion for
  egocentric assistant.
\newblock In {\em ECCV}, 2022.

\bibitem{damen2022rescaling}
Dima Damen, Hazel Doughty, Giovanni~Maria Farinella, Antonino Furnari,
  Evangelos Kazakos, Jian Ma, Davide Moltisanti, Jonathan Munro, Toby Perrett,
  Will Price, et~al.
\newblock Rescaling egocentric vision: Collection, pipeline and challenges for
  epic-kitchens-100.
\newblock {\em IJCV}, 130(1):33--55, 2022.

\bibitem{sigurdsson2018charades}
Gunnar~A Sigurdsson, Abhinav Gupta, Cordelia Schmid, Ali Farhadi, and Karteek
  Alahari.
\newblock Charades-ego: A large-scale dataset of paired third and first person
  videos.
\newblock {\em arXiv preprint arXiv:1804.09626}, 2018.

\bibitem{li2015delving}
Yin Li, Zhefan Ye, and James~M Rehg.
\newblock Delving into egocentric actions.
\newblock In {\em CVPR}, pages 287--295, 2015.

\bibitem{ma2022vip}
Yecheng~Jason Ma, Shagun Sodhani, Dinesh Jayaraman, Osbert Bastani, Vikash
  Kumar, and Amy Zhang.
\newblock Vip: Towards universal visual reward and representation via
  value-implicit pre-training.
\newblock {\em arXiv preprint arXiv:2210.00030}, 2022.

\bibitem{NEURIPS2020_1457c0d6}
Tom Brown, Benjamin Mann, Nick Ryder, Melanie Subbiah, Jared~D Kaplan, Prafulla
  Dhariwal, Arvind Neelakantan, Pranav Shyam, Girish Sastry, Amanda Askell,
  Sandhini Agarwal, Ariel Herbert-Voss, Gretchen Krueger, Tom Henighan, Rewon
  Child, Aditya Ramesh, Daniel Ziegler, Jeffrey Wu, Clemens Winter, Chris
  Hesse, Mark Chen, Eric Sigler, Mateusz Litwin, Scott Gray, Benjamin Chess,
  Jack Clark, Christopher Berner, Sam McCandlish, Alec Radford, Ilya Sutskever,
  and Dario Amodei.
\newblock Language models are few-shot learners.
\newblock In H.~Larochelle, M.~Ranzato, R.~Hadsell, M.F. Balcan, and H.~Lin,
  editors, {\em Advances in Neural Information Processing Systems}, volume~33,
  pages 1877--1901. Curran Associates, Inc., 2020.

\bibitem{visualchatgpt}
Chenfei Wu, Shengming Yin, Weizhen Qi, Xiaodong Wang, Zecheng Tang, and Nan
  Duan.
\newblock Visual chatgpt: Talking, drawing and editing with visual foundation
  models.
\newblock {\em CoRR}, abs/2303.04671, 2023.

\bibitem{yang2023mm}
Zhengyuan Yang, Linjie Li, Jianfeng Wang, Kevin Lin, Ehsan Azarnasab, Faisal
  Ahmed, Zicheng Liu, Ce~Liu, Michael Zeng, and Lijuan Wang.
\newblock Mm-react: Prompting chatgpt for multimodal reasoning and action.
\newblock {\em arXiv preprint arXiv:2303.11381}, 2023.

\bibitem{hugginggpt}
Yongliang Shen, Kaitao Song, Xu~Tan, Dongsheng Li, Weiming Lu, and Yueting
  Zhuang.
\newblock Hugginggpt: Solving {AI} tasks with chatgpt and its friends in
  huggingface.
\newblock {\em CoRR}, abs/2303.17580, 2023.

\bibitem{minigpt4}
Deyao Zhu, Jun Chen, Xiaoqian Shen, Xiang Li, and Mohamed Elhoseiny.
\newblock Minigpt-4: Enhancing vision-language understanding with advanced
  large language models, 2023.

\bibitem{llava}
Haotian Liu, Chunyuan Li, Qingyang Wu, and Yong~Jae Lee.
\newblock Visual instruction tuning.
\newblock {\em CoRR}, abs/2304.08485, 2023.

\bibitem{li2023videochat}
KunChang Li, Yinan He, Yi~Wang, Yizhuo Li, Wenhai Wang, Ping Luo, Yali Wang,
  Limin Wang, and Yu~Qiao.
\newblock Videochat: Chat-centric video understanding.
\newblock {\em arXiv preprint arXiv:2305.06355}, 2023.

\bibitem{ye2023mplug}
Qinghao Ye, Haiyang Xu, Guohai Xu, Jiabo Ye, Ming Yan, Yiyang Zhou, Junyang
  Wang, Anwen Hu, Pengcheng Shi, Yaya Shi, et~al.
\newblock mplug-owl: Modularization empowers large language models with
  multimodality.
\newblock {\em arXiv preprint arXiv:2304.14178}, 2023.

\bibitem{chen2023x}
Feilong Chen, Minglun Han, Haozhi Zhao, Qingyang Zhang, Jing Shi, Shuang Xu,
  and Bo~Xu.
\newblock X-llm: Bootstrapping advanced large language models by treating
  multi-modalities as foreign languages.
\newblock {\em arXiv preprint arXiv:2305.04160}, 2023.

\bibitem{palm-e}
Danny Driess, Fei Xia, Mehdi S.~M. Sajjadi, Corey Lynch, Aakanksha Chowdhery,
  Brian Ichter, Ayzaan Wahid, Jonathan Tompson, Quan Vuong, Tianhe Yu, Wenlong
  Huang, Yevgen Chebotar, Pierre Sermanet, Daniel Duckworth, Sergey Levine,
  Vincent Vanhoucke, Karol Hausman, Marc Toussaint, Klaus Greff, Andy Zeng,
  Igor Mordatch, and Pete Florence.
\newblock Palm-e: An embodied multimodal language model.
\newblock {\em CoRR}, abs/2303.03378, 2023.

\bibitem{ahn2022can}
Michael Ahn, Anthony Brohan, Noah Brown, Yevgen Chebotar, Omar Cortes, Byron
  David, Chelsea Finn, Keerthana Gopalakrishnan, Karol Hausman, Alex Herzog,
  et~al.
\newblock Do as i can, not as i say: Grounding language in robotic affordances.
\newblock {\em arXiv preprint arXiv:2204.01691}, 2022.

\bibitem{touvron2023llama}
Hugo Touvron, Thibaut Lavril, Gautier Izacard, Xavier Martinet, Marie-Anne
  Lachaux, Timoth{\'e}e Lacroix, Baptiste Rozi{\`e}re, Naman Goyal, Eric
  Hambro, Faisal Azhar, et~al.
\newblock Llama: Open and efficient foundation language models.
\newblock {\em arXiv preprint arXiv:2302.13971}, 2023.

\bibitem{coco}
Tsung{-}Yi Lin, Michael Maire, Serge~J. Belongie, James Hays, Pietro Perona,
  Deva Ramanan, Piotr Doll{\'{a}}r, and C.~Lawrence Zitnick.
\newblock Microsoft {COCO:} common objects in context.
\newblock In David~J. Fleet, Tom{\'{a}}s Pajdla, Bernt Schiele, and Tinne
  Tuytelaars, editors, {\em {ECCV}}, volume 8693, pages 740--755, 2014.

\bibitem{cc3m}
Piyush Sharma, Nan Ding, Sebastian Goodman, and Radu Soricut.
\newblock Conceptual captions: A cleaned, hypernymed, image alt-text dataset
  for automatic image captioning.
\newblock In {\em Proceedings of the 56th Annual Meeting of the Association for
  Computational Linguistics (Volume 1: Long Papers)}, pages 2556--2565, 2018.

\bibitem{liu2023visual}
Haotian Liu, Chunyuan Li, Qingyang Wu, and Yong~Jae Lee.
\newblock Visual instruction tuning.
\newblock {\em arXiv preprint arXiv:2304.08485}, 2023.

\bibitem{singh20193d}
Rahul~Dev Singh, Ajay Mittal, and Rajesh~K Bhatia.
\newblock 3d convolutional neural network for object recognition: a review.
\newblock {\em Multimedia Tools and Applications}, 78:15951--15995, 2019.

\bibitem{he2016deep}
Kaiming He, Xiangyu Zhang, Shaoqing Ren, and Jian Sun.
\newblock Deep residual learning for image recognition.
\newblock In {\em Proceedings of the IEEE conference on computer vision and
  pattern recognition}, pages 770--778, 2016.

\bibitem{deng2009imagenet}
Jia Deng, Wei Dong, Richard Socher, Li-Jia Li, Kai Li, and Li~Fei-Fei.
\newblock Imagenet: A large-scale hierarchical image database.
\newblock In {\em 2009 IEEE conference on computer vision and pattern
  recognition}, pages 248--255. Ieee, 2009.

\bibitem{riedmiller2014multi}
Martin Riedmiller and A~Lernen.
\newblock Multi layer perceptron.
\newblock {\em Machine Learning Lab Special Lecture, University of Freiburg},
  pages 7--24, 2014.

\bibitem{eva}
Yuxin Fang, Wen Wang, Binhui Xie, Quan Sun, Ledell Wu, Xinggang Wang, Tiejun
  Huang, Xinlong Wang, and Yue Cao.
\newblock Eva: Exploring the limits of masked visual representation learning at
  scale.
\newblock {\em arXiv preprint arXiv:2211.07636}, 2022.

\bibitem{gpt4_ins}
Baolin Peng, Chunyuan Li, Pengcheng He, Michel Galley, and Jianfeng Gao.
\newblock Instruction tuning with gpt-4, 2023.

\bibitem{dosovitskiy2020image}
Alexey Dosovitskiy, Lucas Beyer, Alexander Kolesnikov, Dirk Weissenborn,
  Xiaohua Zhai, Thomas Unterthiner, Mostafa Dehghani, Matthias Minderer, Georg
  Heigold, Sylvain Gelly, et~al.
\newblock An image is worth 16x16 words: Transformers for image recognition at
  scale.
\newblock In {\em ICLR}, 2020.

\bibitem{NEURIPS2022_31fb284a}
Kevin~Qinghong Lin, Jinpeng Wang, Mattia Soldan, Michael Wray, Rui Yan, Eric~Z.
  XU, Difei Gao, Rong-Cheng Tu, Wenzhe Zhao, Weijie Kong, Chengfei Cai, WANG
  HongFa, Dima Damen, Bernard Ghanem, Wei Liu, and Mike~Zheng Shou.
\newblock Egocentric video-language pretraining.
\newblock In S.~Koyejo, S.~Mohamed, A.~Agarwal, D.~Belgrave, K.~Cho, and A.~Oh,
  editors, {\em Advances in Neural Information Processing Systems}, volume~35,
  pages 7575--7586. Curran Associates, Inc., 2022.

\bibitem{openai_chatgpt_2023}
OpenAI.
\newblock Chatgpt (mar 14 version) [large language model], 2023.

\bibitem{clip}
Alec Radford, Jong~Wook Kim, Chris Hallacy, Aditya Ramesh, Gabriel Goh,
  Sandhini Agarwal, Girish Sastry, Amanda Askell, Pamela Mishkin, Jack Clark,
  et~al.
\newblock Learning transferable visual models from natural language
  supervision.
\newblock {\em arXiv preprint arXiv:2103.00020}, 2021.

\bibitem{puig2018virtualhome}
Xavier Puig, Kevin Ra, Marko Boben, Jiaman Li, Tingwu Wang, Sanja Fidler, and
  Antonio Torralba.
\newblock Virtualhome: Simulating household activities via programs.
\newblock In {\em Proceedings of the IEEE Conference on Computer Vision and
  Pattern Recognition}, pages 8494--8502, 2018.

\bibitem{adamw}
Ilya Loshchilov and Frank Hutter.
\newblock Decoupled weight decay regularization.
\newblock {\em arXiv preprint arXiv:1711.05101}, 2017.

\end{thebibliography}
\bibliographystyle{unsrt}
}

\newpage
\appendix

\section{Implementation details}
\subsection{Hyper-parameters}
We use the same set of training hyper-parameters for all models during vision-language pre-training. We employ the AdamW optimizer~\cite{adamw} with $\beta_1=0.9$, $\beta_2=0.98$, and a weight decay of 0.05. We also utilize a cosine learning rate decay with a peak learning rate of $2\times10^{-5}$ and a linear warm-up with warm-up ratio $5\times10^{-2}$. Our training data consists of images of size 224$\times$224 that are augmented with random resized cropping and horizontal flipping. The maximize sequence length is set as 256.

\subsection{Downstream policy learning }
We adopt imitation learning as the method of policy learning in low level control tasks, which leverages demonstration data provided by an expert to learn the desired behavior. This technique has found applications in various domains, such as robotics, autonomous driving, and game playing. We provide each task 25 demonstrations, which are trajectories of observations and actions performed by an expert in the given task, and test the performance with 25 demonstrations and only 10 demonstrations respectively. The goal of imitation learning is to learn a policy, denoted as $\pi$, that maps the agent's observations to appropriate actions. The learned policy should be able to imitate the expert's behavior accurately. Speciffically, we use behavioral cloning to learn the downstream policy, which trains a supervised learning model to predict actions given states based on the expert demonstrations, and the loss function is shown as Equation \ref{eq:imitation}

\begin{equation}
   L(\theta) = \sum [\pi_{\theta}(a|s) \log P^*(a|s)] 
   \label{eq:imitation}
\end{equation}

Here, $\theta$ represents the parameters of the policy model, $\pi_{\theta}(a|s)$ denotes the predicted action probability distribution given a state $s$, and $P^*(a|s)$ represents the ground truth action probability distribution derived from the expert demonstrations.

For Franka-kitchen tasks, the length of demonstration is 50, which contains 50 state-action pairs. For Meta-World tasks, the length of demonstration is 500, which contains 500 state-action pairs.

\section{More demos of \model}
\subsection{Visual Captioning}
We assessed EmbodiedGPT on numerous visual captioning tasks spanning a range of embodied AI benchmarks. As shown in Figure \ref{fig:image_caption}, the model displayed an exceptional ability to accurately describe objects, characters, and spatial relationships relevant to embodied AI tasks. Furthermore, EmbodiedGPT exhibited robust zero-shot learning capabilities, evidenced by its strong performance across multiple benchmarks without the need for task-specific fine-tuning.

\begin{figure}[t]
    \centering
    \begin{subfigure}[b]{1.0\textwidth}
        \centering        \includegraphics[width=\textwidth]{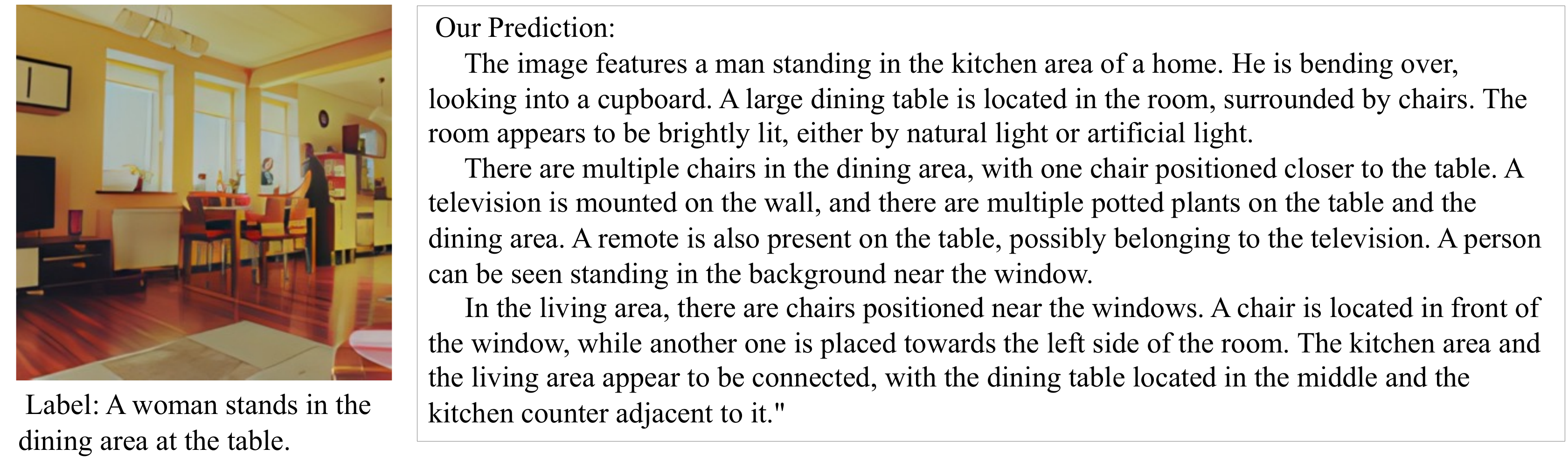}
        \caption{Image caption in COCO dataset.}
        \label{fig:caption_ex1}
    \end{subfigure}
    \hfill
 %    \begin{subfigure}[b]{0.98\textwidth}
 %        \centering
 % \includegraphics[width=\textwidth]{figure/caption_example_2.pdf}
 %        \caption{Image caption in living room in COCO dataset.}
 %        \label{fig:caption_ex2}
 %    \end{subfigure}
        \begin{subfigure}[b]{1.0\textwidth}
        \centering
 \includegraphics[width=\textwidth]{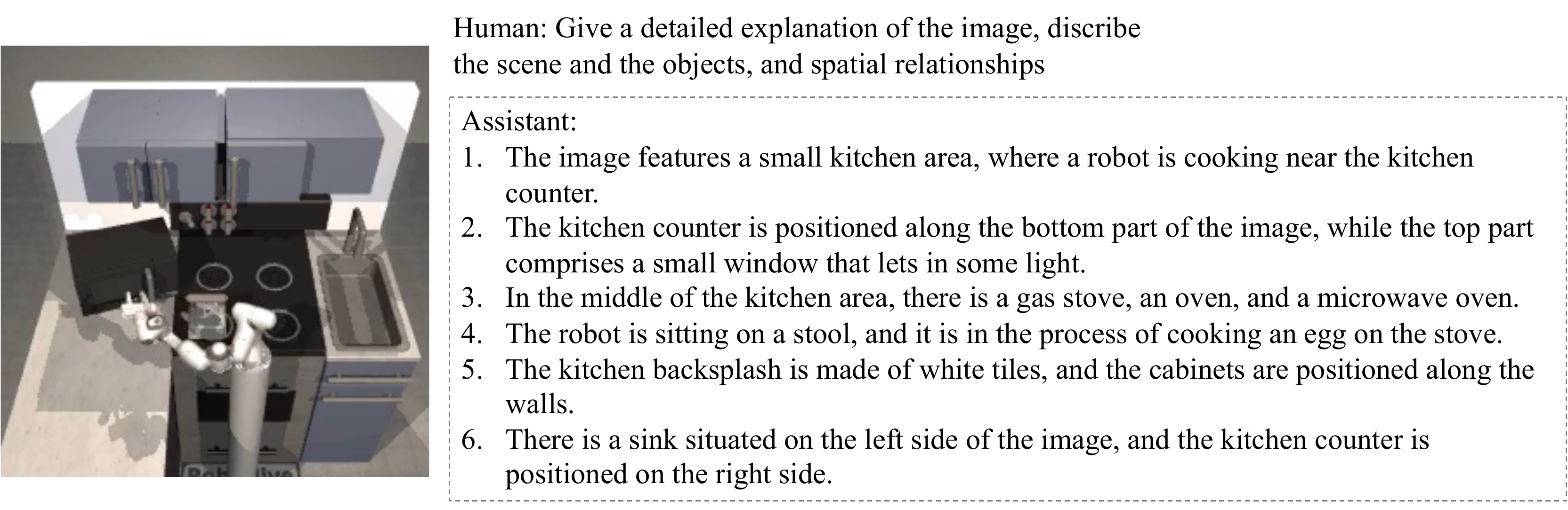}
        \caption{Embodied image caption in Franka Kitchen benchmark.}
        \label{fig:caption_ex3}
    \end{subfigure}
         \begin{subfigure}[b]{1.0\textwidth}
        \centering
\includegraphics[width=\textwidth]{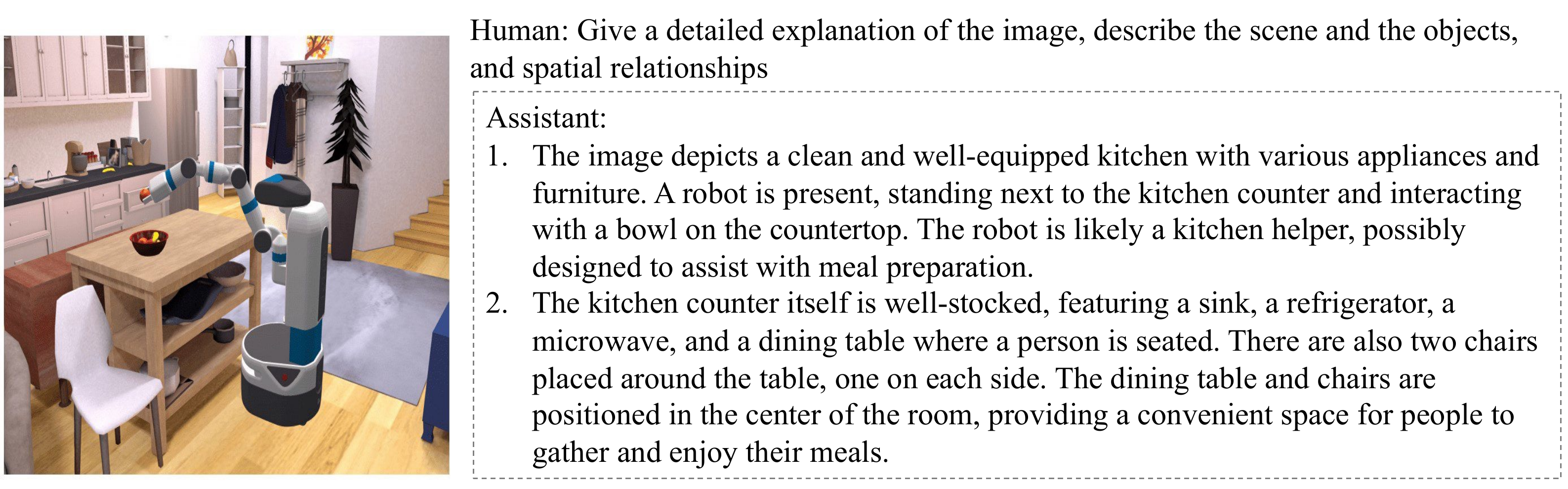}
        \caption{Embodied image caption in Habitat2.0 benchmark.}
        \label{fig:caption_ex4}
    \end{subfigure}
    \caption{Generation results in image caption tasks.}
    \label{fig:image_caption}
\end{figure}
% $$ $$
% $$ $$
% $$ $$

\subsection{Embodied Planning with image input}
\textbf{Embodied Planning for Concrete Tasks (image input)}:
In the context of concrete task planning, such as making a cup of coffee, \model effectively utilized visual information to pinpoint the required objects and their positional relationships within the environment. The model produced coherent, multi-step plans, taking into consideration both the task requirements and environmental constraints. This capability demonstrates the model’s potential to facilitate real-world applications and tackle complex planning challenges within the realm of embodied AI.
\begin{figure}
    \centering
    % \begin{subfigure}[b]{0.98\textwidth}
    %     \centering        \includegraphics[width=\textwidth]{figure/real_caption.pdf}
    %     \caption{Recognize the scene in real world.}
    %     \label{fig:caption_ex1}
    % \end{subfigure}
    % \hfill
 %    \begin{subfigure}[b]{0.98\textwidth}
 %        \centering
 % \includegraphics[width=\textwidth]{figure/caption_example_2.pdf}
 %        \caption{Image caption in living room in COCO dataset.}
 %        \label{fig:caption_ex2}
 %    \end{subfigure}
        % \begin{subfigure}[b]{0.98\textwidth}
        % \centering
 \includegraphics[width=\textwidth]{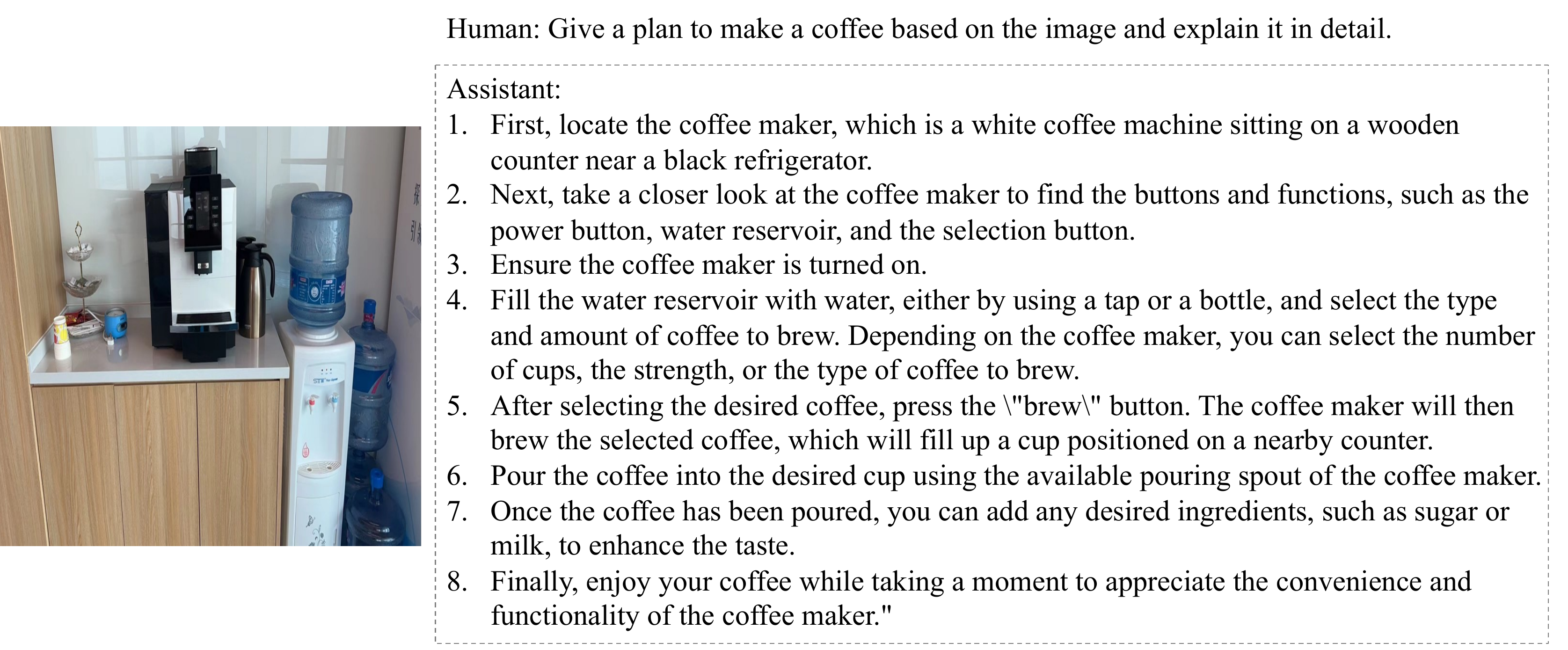}
        \caption{Embodied planning of in real-world scenarios.}
        \label{fig:caption_ex3}
\end{figure}

\textbf{Embodied Planning for Abstract Tasks}:
For abstract task scenarios, EmbodiedGPT adeptly combined visual observations with abstract concepts to generate concrete sub-task descriptions. For instance, when given the abstract prompt of feeling hot, the model identified pertinent objects in the environment (e.g., a fan) and suggested a practical solution (e.g., turning on the fan). Subsequently, the model generated a detailed plan to accomplish the identified sub-tasks, highlighting its adaptability across diverse problem-solving contexts.

\begin{figure}
    \centering
    % \begin{subfigure}[b]{0.98\textwidth}
    %     \centering        \includegraphics[width=\textwidth]{figure/real_caption.pdf}
    %     \caption{Recognize the scene in real world.}
    %     \label{fig:caption_ex1}
    % \end{subfigure}
    % \hfill
 %    \begin{subfigure}[b]{0.98\textwidth}
 %        \centering
 % \includegraphics[width=\textwidth]{figure/caption_example_2.pdf}
 %        \caption{Image caption in living room in COCO dataset.}
 %        \label{fig:caption_ex2}
 %    \end{subfigure}
            \begin{subfigure}[b]{0.98\textwidth}
        \centering
 \includegraphics[width=\textwidth]{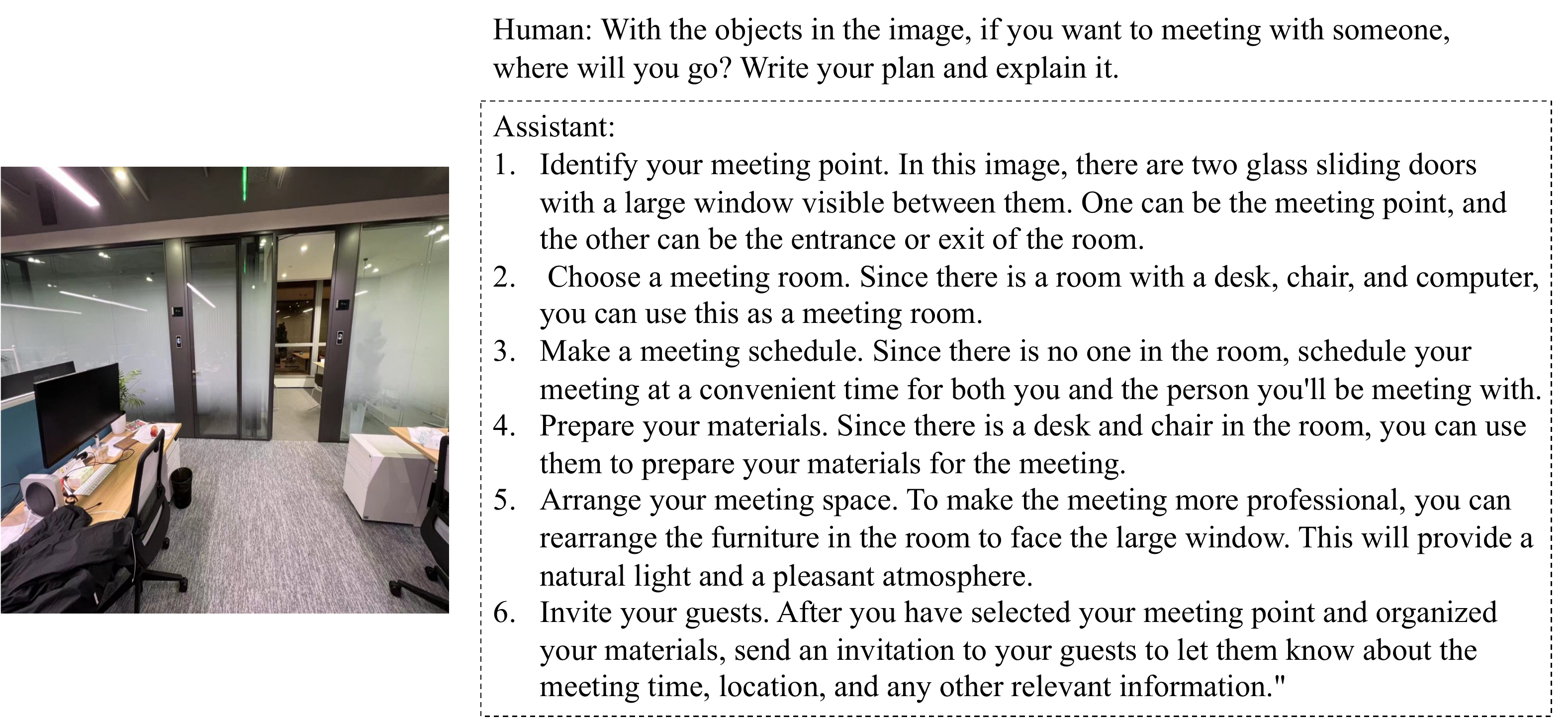}
        \caption{Embodied planning for abstract tasks [meeting].}
        \label{fig:caption_ex3}
    \end{subfigure}
                \begin{subfigure}[b]{0.98\textwidth}
        \centering
 \includegraphics[width=\textwidth]{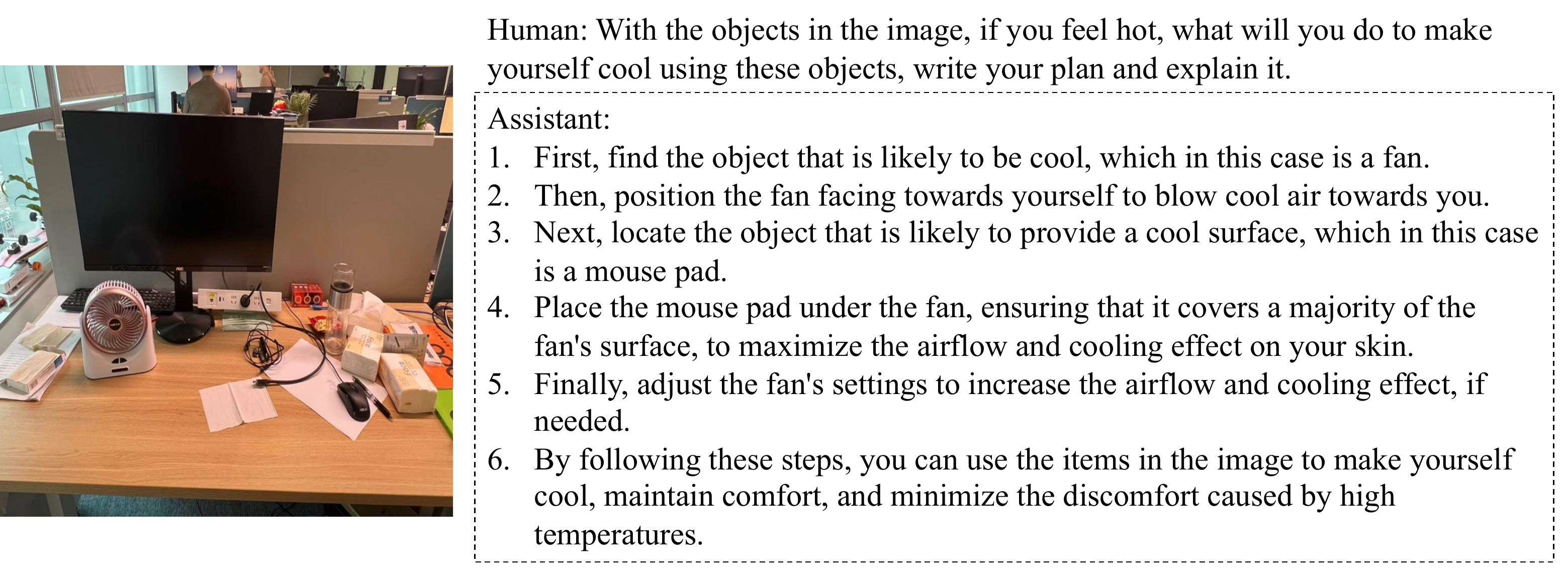}
        \caption{Embodied planning for abstract tasks[feel hot].}
        \label{fig:caption_ex3}
    \end{subfigure}
    \caption{Embodied planning for abstract tasks.}
    \label{fig:image_caption}
\end{figure}

\section{Evaluation metric and scoring criteria for user study}

We show the Table \ref{tab:scoring criteria} to outline the scoring criteria for a user study incorporating the above five evaluation metrics:

\begin{table}[]
    \centering
\begin{tabular}{p{3cm}|p{10cm}}
\toprule
\textbf{Evaluation Metric} & \textbf{Explanation} \\
\hline
Object Recognition Accuracy & This metric measures the ability of a system to accurately identify objects from images or videos. A higher accuracy indicates that the system can correctly recognize the objects present in the given visual data. \\
\hline
Spatial Relationship Understanding & Spatial relationship understanding refers to the system's capability to accurately discern the spatial relationships between objects in a scene. It evaluates whether the system can determine the relative positions, orientations, distances, and other spatial attributes of objects with precision. \\
\hline
Level of Redundancy in the Answer & The level of redundancy in the answer assesses the amount of unnecessary or repetitive information present in the system's response. Lower redundancy indicates that the system provides concise and non-repetitive answers, which is generally preferred as it reduces verbosity and improves clarity. \\
\hline
Reasonability of the Planning & The reasonability of the planning metric gauges the logical coherence and appropriateness of the system's planning process. It examines whether the system's generated plans are sensible and align with the given goals or objectives. \\
\hline
Executability of the Planning & This metric evaluates the feasibility and practicality of the system's generated plans. It assesses whether the plans can be executed successfully in the real world. \\
\bottomrule
\end{tabular}
\caption{Explanation of Evaluation Metric}
    \label{tab:scoring criteria}
\end{table}

\section{Insight about the prompt designing for multi-modal large model}
Through extensive experiments evaluating multi-modal large models in question answering tasks, we have identified a limitation in their ability to effectively utilize visual information. These models tend to rely heavily on textual inputs and may not fully exploit the visual cues present in the question-answering process. In order to address this limitation and enhance their performance, we aimed to investigate the impact of incorporating additional prompts that guide the model's attention towards the visual content.

To achieve this goal, we propose a straightforward yet highly effective approach: incorporating additional prompting into the model's input. This involves introducing specific prompts such as "in the scene shown in this image/video" or allowing the model to describe the image/video as part of a multi-turn dialogue. By including these prompts, we aim to explicitly direct the model's focus towards the visual information available and encourage it to utilize this information when generating answers.
Our experiments have yielded promising results. The introduction of additional prompts has significantly improved the model's ability to leverage visual information and provide accurate answers based on the visual content. By explicitly referencing the scene depicted in the image or video, the model's attention is directed towards the relevant visual features, leading to a more comprehensive integration of visual and textual information. Consequently, the model's reasoning ability is enhanced, resulting in more precise and contextually grounded answers.

% \begin{figure}[t]
% % \begin{minipage}[t]{0.59\linewidth}
% \centering
% \includegraphics[width=1.0\linewidth]{figure/multi_round.pdf}
% \caption{Multi-turn dialogue and  embodied planning with input video. }
% \label{fig:Overall framework}
% \end{figure}

\begin{figure}
    \centering
    \begin{subfigure}[b]{0.98\textwidth}
        \centering        \includegraphics[width=\textwidth]{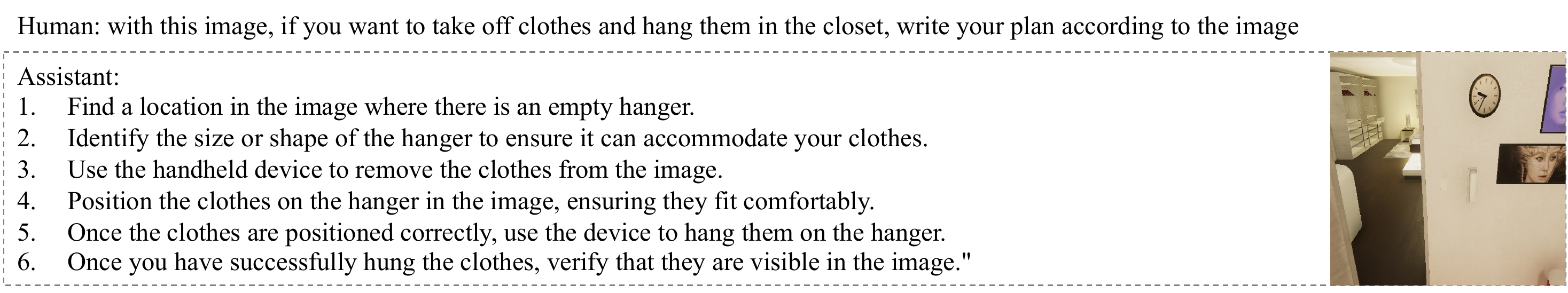}
        \caption{Ask EmbodiedGPT write the plan directly.}
        \label{fig:y equals x}
    \end{subfigure}
    \hfill
    \begin{subfigure}[b]{0.98\textwidth}
        \centering
 \includegraphics[width=\textwidth]{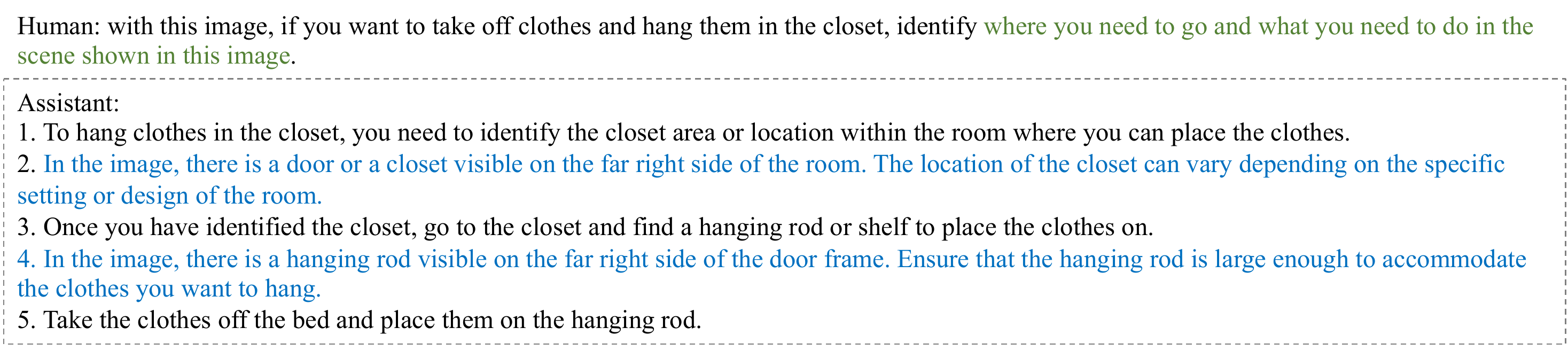}
        \caption{Ask EmbodiedGPT write the plan directly with \textit{\textbf{image-related chain-of-thought}}.}
        \label{fig:three sin x}
    \end{subfigure}
    \caption{Performance Comparison with different types prompt.}
    \label{fig:three graphs}
\end{figure}

% This paper presents \model, an end-to-end multi-modal foundation model for embodied AI that enables robots to generate step by step planning and execute action sequences. Our work includes the crafting of a large-scale visual embodied planning dataset, \dataset, which contains carefully selected high-quality video and text planning pairs. We introduce a cost-effective training approach for high-quality plan generation by adapting a 7B language model to the \dataset using prefix adapters. Additionally, we design a paradigm for extracting task-relevant features from LLM-generated planning queries, thereby forming a closed loop between high-level planning and low-level control. Through extensive experiments on various embodied tasks, including embodied planning, embodied control, visual captioning, and visual question answering, \model shows excellent performance and its potential to become a solid foundation model for embodied AI community.

% \bibliographystyle{name}
% \bibliography{main}

%%%%%%%%%%%%%%%%%%%%%%%%%%%%%%%%%%%%%%%%%%%%%%%%%%%%%%%%%%%%

\end{document}